\documentclass[journal]{IEEEtran}
\usepackage{amsmath}
\usepackage{multirow,booktabs}
\usepackage{threeparttable}
\usepackage{makecell}
\usepackage{moreverb}
\usepackage{bbm}
\usepackage{cite}
\usepackage{bm}
\usepackage{mathtools}   
\usepackage{setspace}
\usepackage[justification=centering]{caption}
\usepackage{graphicx}
\usepackage{float}
\usepackage{caption}
\usepackage{subcaption}
\usepackage{tabularx}
\usepackage{booktabs}
\usepackage{multirow} 
\usepackage{xcolor}

\usepackage{float}

\usepackage[colorlinks,bookmarksopen,bookmarksnumbered,citecolor=red]{hyperref}
\DeclareSymbolFont{ugmL}{OMX}{mdugm}{m}{n}
\SetSymbolFont{ugmL}{bold}{OMX}{mdugm}{b}{n}
\DeclareMathAccent{\wideparen}{\mathord}{ugmL}{"F3}

\begin{document}
\captionsetup{labelformat=simple,labelsep=period}

\title{A Novel Hybrid Time-Varying Graph Neural Network For Traffic Flow Forecasting}

\author{Ben-Ao Dai,~
        Bao-Lin~Ye,~
        Lingxi~Li~
\thanks{This work was supported in part by Zhejiang Provincial Natural Science Foundation of China under Grant No.LTGS23F030002; and by the Jiaxing Public Welfare Research Program No.2023AY11034; and by the National Natural Science Foundation of China under Grant No. 61603154; and by the Open Research Project of the State Key Laboratory of Industrial Control Technology, Zhejiang University, China (No. ICT2022B52).
}
\thanks{Ben-Ao Dai and Bao-Lin Ye are with the School of Information Science and Engineering, Jiaxing University, Jiaxing, Zhejiang, 314001, China; and also with the School of Information Science and Engineering, Zhejiang Sci-tech University, Xiasha Campus, Hangzhou, Zhejiang, 310018, China.}
%
%
\thanks{Lingxi Li is with the Elmore Family School of Electrical and Computer Engineering, Purdue University, Indianapolis, IN 46202, USA (email:lingxili@purdue.edu).}
\thanks{Corresponding author: Bao-Lin Ye (yebaolin@zjxu.edu.cn) }
}
\maketitle
\begin{abstract}
Real-time and precise traffic flow prediction is vital for the efficiency of intelligent transportation systems. Traditional methods often employ graph neural networks (GNNs) with pre-defined graphs to describe spatial correlations among traffic nodes in urban road networks. However, these pre-defined graphs are limited by existing knowledge and graph generation methodologies, offering an incomplete picture of spatial correlations. 
While time-varying graphs based on data-driven learning have attempted to address these limitations, they still struggle with adequately capturing the inherent spatial correlations in traffic data. 
Moreover, most current methods for capturing dynamic temporal correlations rely on a unified calculation scheme using a temporal multi-head self-attention mechanism, which at some level might leads to inaccuracies. 
%
%
In order to overcome these challenges, we have proposed a novel hybrid time-varying graph neural network (HTVGNN) for traffic flow prediction. Firstly, a novel enhanced temporal perception multi-head self-attention mechanism based on time-varying mask enhancement was reported to more accurately model the dynamic temporal dependencies among distinct traffic nodes in the traffic network. Secondly, we have proposed a novel graph learning strategy to concurrently learn both static and dynamic spatial associations between different traffic nodes in road networks. Meanwhile, in order to enhance the learning ability of time-varying graphs, a coupled graph learning mechanism was designed to couple the graphs learned at each time step. Finally, the effectiveness of the proposed method HTVGNN was demonstrated with four real data sets. Simulation results revealed that HTVGNN achieves superior prediction accuracy compared to the state of the art spatio-temporal graph neural network models. Additionally, the ablation experiment verifies that the coupled graph learning mechanism can effectively improve the long-term prediction performance of HTVGNN. 
%
%
%
%
%
%

\end{abstract}


\IEEEpeerreviewmaketitle

\section{Introduction}

\IEEEPARstart{T}{he} rapid acceleration of urbanization has led to a sharp increase in car ownership. This rise in vehicles has intensified the challenge of enhancing road network efficiency to alleviate traffic congestion~\cite{ref-19}. Traffic flow prediction, offering future traffic state information based on real-time data, is pivotal for advancing intelligent traffic management and control.

Early research primarily utilized statistical methods for constructing traffic flow prediction models. For example, historical average (HA) \cite{ref-18} \cite{ref-17}, autoregressive moving average model (ARIMA) \cite{ref-16} \cite{ref-15}, vector autoregressive model (VAR) \cite{ref-14}, etc. These statistics based methods assume linear dependencies in time-series data. However, the complex and variable nature of traffic flow leads to nonlinear dependencies, resulting in these methods' subpar performance in practical forecasting tasks. After that, traditional machine learning-based methods (e.g. support vector regression (SVR) \cite{ref-13} and k-nearest neighbor (KNN) \cite{ref-12}) were reported to overcome the drawbacks of statistics based methods. Nevertheless, such methods' reliance on handcrafted features limits their effectiveness in leveraging the growing abundance of large-scale traffic flow data. After that, long short-term memory (LSTM) and their variants gated recurrent units (GRU) \cite{ref-11} were used to model the temporal dependencies of traffic flow time-series data, and  convolutional neural networks (CNN) \cite{ref-10} and graph convolutional networks (GCN) \cite{ref-9} \cite{ref-8} \cite{ref-7} were employed to capture the spatial dependencies of different traffic nodes.

In recent years, since the core of traffic flow prediction is to accurately model the complex and nonlinear spatial-temporal dependencies of dynamic traffic flows of different traffic nodes, graph convolutional network (GCN) and Transformer \cite{ref-3-1} are two widely used core components of many representative traffic flow prediction models~\cite{ref--6,ref-5,ref-4,ref-4-1}.
While such spatial-temporal modeling approaches have yielded promising results in traffic flow prediction, accurately capturing the complex and dynamic spatial-temporal correlations within road networks remains challenging for the following reasons.
\begin{itemize}
    \item Initially developed for natural language processing, the Transformer model\cite{ref-3-1} excels in machine translation due to its strong ability to model long-term contextual features. Its Encoder-Decoder architecture notably improves model generalization. Central to this architecture is the multi-head self-attention mechanism. However, this traditional mechanism struggles to effectively model dynamic temporal correlations in traffic flow, often leading to errors in weight calculations~\cite{ref-4}. Additionally, integrating periodic features into this mechanism could potentially enhance its capacity to perceive time-related variations in traffic data.
    \item It is challenging to capture the spatial correlations between road network nodes without prior knowledge. Although some works try to create spatial structure graphs based on existing knowledge~\cite{ref-3, ref-3-2}, these graphs often inaccurately represent spatial correlations due to the limitations of their underlying assumptions and methods. Some other works~\cite{ref-3-3, ref-2} have made strides in adaptive spatial modeling without fixed graphs, but they still predominantly use a global adaptive graph across all time steps, restricting their learning potential.
    \item Additionally, a major challenge lies in the static nature of spatial correlations in both adaptive and predefined graphs, which doesn't align with the dynamic and ever-changing topology of traffic networks. Accurately modeling these dynamic spatial correlations is crucial. Existing models like Liu et al.'s STGHTN \cite{ref-5} and Zhu et al.'s CorrSTN \cite{ref-4-1} use multi-head attention mechanisms for dynamic modeling-the former for adjusting features dynamically and the latter for modifying graphs before cluster convolution. However, these approaches typically entail high computational complexity due to their reliance on spatial multi-head attention.
\end{itemize}

To tackle the identified challenges, this paper proposed a novel hybrid time-varying graph neural network for traffic flow prediction. A novel enhanced temporal perception multi-head self-attention mechanism enhanced by time-varying masks was presented to boost the computational accuracy and temporal sensitivity of attention models. An efficient coupled time-varying graph convolutional recurrent network was designed to capture the spatio-temporal dynamics of traffic flows. This network leverages time-varying graphs for each time step, enhancing its ability to learn spatial dependencies and long-term trends in road network traffic. By integrating time-varying graphs across different time steps, our model effectively incorporates both short-term and long-term traffic patterns. Furthermore, we proposed a dynamic graph convolution approach, which utilized a combined topology and semantic matrix as a mask, enabling more precise modeling of the dynamic spatial correlations in traffic networks.


The following are three key contributions of our work in traffic flow prediction:
\begin{itemize}
    \item
    A novel temporal perception multi-head self-attention mechanism enhanced by a time-varying mask, diverging from traditional approaches is proposed. This mechanism dynamically adjusts attention calculations based on the temporal characteristics of input data, enabling more accurate capture of temporal dependencies among traffic nodes. The time-varying mask consists of three components: a static mask embedding for correcting the multi-head self-attention mechanism, and two dynamic mask embeddings to augment the time-awareness of the mechanism.
    \item
    A new graph learning strategy is reported to simultaneously address static and dynamic spatial correlations in traffic networks. To enhance learning of static spatial correlations, we employ both global and local time-varying embeddings, facilitating the understanding of spatial relationships at various time steps. Additionally, a coupling graph learning mechanism is implemented to integrate graphs learned at different time steps. For more precise capture of dynamic spatial correlations, we define a mask matrix, derived from topology and semantic matrices, to refine the dynamic graphs produced by our strategy.
    \item
    Simulation on four real-world datasets is made to evaluate our proposed HTVGNN method. The results demonstrate the effectiveness of HTVGNN, showing higher prediction accuracy than the latest space-time graph neural network models. Furthermore, an ablation study confirms that the integrated coupled graph learning mechanism significantly enhances the model's long-term prediction capabilities.
\end{itemize}

%
%
\section{Related Work}
%
%
\subsection{Traditional Traffic Forecasting
Methods}
Traffic flow prediction plays a pivotal role in urban intelligent transportation, enabling the anticipation of future traffic conditions based on real-time data from the traffic network. This information serves as a foundation for scheduling and decision-making by the traffic management department \cite{ref-1}. In earlier studies, researchers predominantly relied on mathematical statistical methods and early machine learning techniques, such as historical average (HA) \cite{ref-8} \cite{ref-9}, autoregressive integrated moving average (ARIMA) \cite{ref-10} \cite{ref-9}, vector autoregressive model (VAR) \cite{ref-3}, etc. However, these models often underperform in actual traffic prediction tasks due to the complex and dynamic nature of traffic flow. With significant advancements in deep learning across various domains like speech recognition and image processing, an increasing number of researchers have started employing deep learning approaches for spatio-temporal traffic flow prediction tasks. Convolutional Neural Networks (CNN) \cite{ref-10}, Long Short-Term Memory (LSTM) networks \cite{ref-11} along with their variant Gate Recurrent Unit (GRU) \cite{ref-11} are widely used and representative deep learning methods that have significantly enhanced the accuracy of traffic flow prediction. Unlike previous methods, deep learning-based approaches can effectively capture spatio-temporal correlations. For instance, Fully Connected Long Short-Term Memory Network (FC-LSTM) \cite{ref--6} combines CNN and LSTM to extract implicit spatio-temporal characteristics present in traffic flow data. Nevertheless, CNNs, originally designed for Euclidean data with regular grid structures such as images and videos, do not accurately process the spatial features of road networks, leading to suboptimal predictive performance in FC-LSTM.
%
%
\subsection{Spatio-temporal graph neural network Traffic Forecasting Methods}
%
%
Traditional traffic research considers traffic flow prediction as a simple time series problem, neglecting the influence of other nodes in the traffic network on the current node. Consequently, the performance of traditional prediction methods is suboptimal. To enhance modeling capabilities, researchers have started investigating effective approaches to capture spatial features within traffic networks. For instance, Conv-LSTM\cite{ref7} employs convolution operations to model spatial features in sequences. However, conventional convolutional neural networks (CNNs) were originally designed for processing Euclidean data with regular grid structures like images and videos, making them ineffective for graph-structured data processing. Graph Convolutional Networks (GCNs) extend traditional convolutions to handle non-Euclidean data and have demonstrated promising results across various graph-based prediction tasks. Spatio-temporal Graph Neural Networks (STGNNs), such as STGCN\cite{ref-3-2}, are increasingly studied in traffic flow prediction research. STGCN combines graph convolutions with time convolutions or recurrent neural networks (RNNs), effectively capturing both spatial and temporal features within road networks. However, its ability to capture spatial features is limited by predefined graph structures used during training. In contrast, models like GraphWaveNet\cite{ref9} and AGCRN\cite{ref-3-3} adaptively learn the graph structure of road networks from data without prior knowledge and achieve comparable prediction performance to models based on predefined graphs.
%
%
\subsection{Transformer}
%
%
The Transformer model was initially proposed in the field of machine translation \cite{ref-3-1} and has since been widely adopted in deep learning domains, including NLP \cite{ref12} and computer vision \cite{ref13}.It features a codec structure implemented with multi-head self-attention (MSA), known for its robust modeling of contextual information. Consequently, MSA has found extensive applications in machine translation tasks, delivering superior performance. For traffic flow prediction, the Transformer model employs a time-based multi-head attention mechanism, mapping data to high-dimensional space through multiple attention groups. This enables simultaneous learning of optimal attention between different features. Researchers have successfully introduced MSA into traffic flow prediction, yielding significant improvements across various aspects. For instance, Yan et al. \cite{ref14} proposed the Traffic Transformer model by first employing a designed coding and embedding block to address data discrepancies across different research fields. Additionally, they enhanced the original encoder-decoder structure of Transformer by introducing two components: global encoder and global-local decoder blocks. These blocks are stacked together to form a deep prediction network with hierarchical characteristics. Wang et al.\cite{ref15}, on the other hand, presented a novel graph neural network layer incorporating position-wise attention mechanism that dynamically aggregates historical traffic flow information from adjacent roads. To better capture temporal features at both local and global levels, they combine RNN with a Transformer layer.

%
%
\begin{figure*}[!htb]
\centering
\includegraphics[width=18cm]{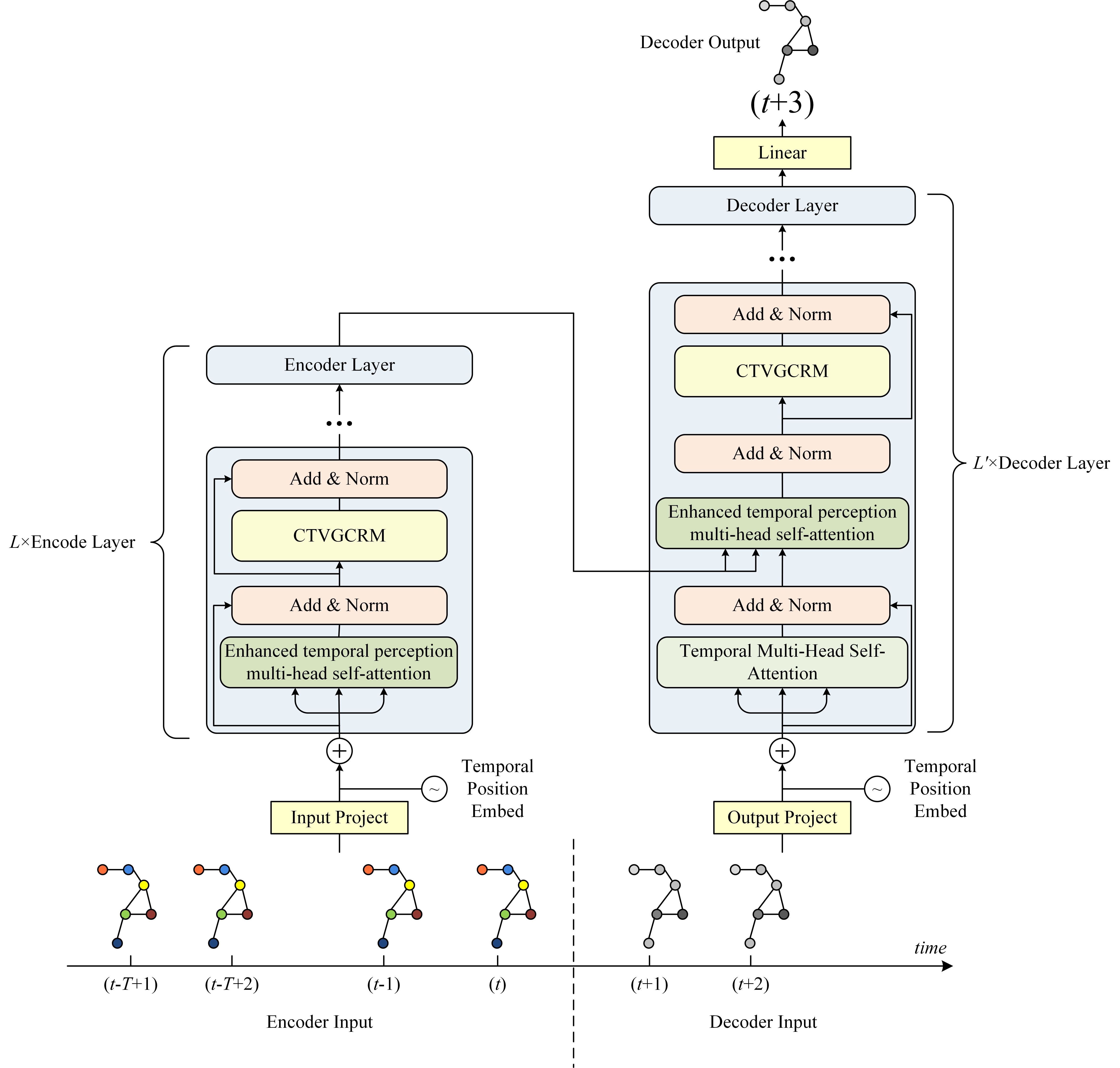}
\caption{The diagram of the overall architecture of the HTVGNN.}
\label{fg:0001}
\end{figure*}
%
%
\begin{figure*}
     \centering
     \begin{subfigure}[b]{0.42\textwidth}
         \centering
         \includegraphics[width=0.9\textwidth]{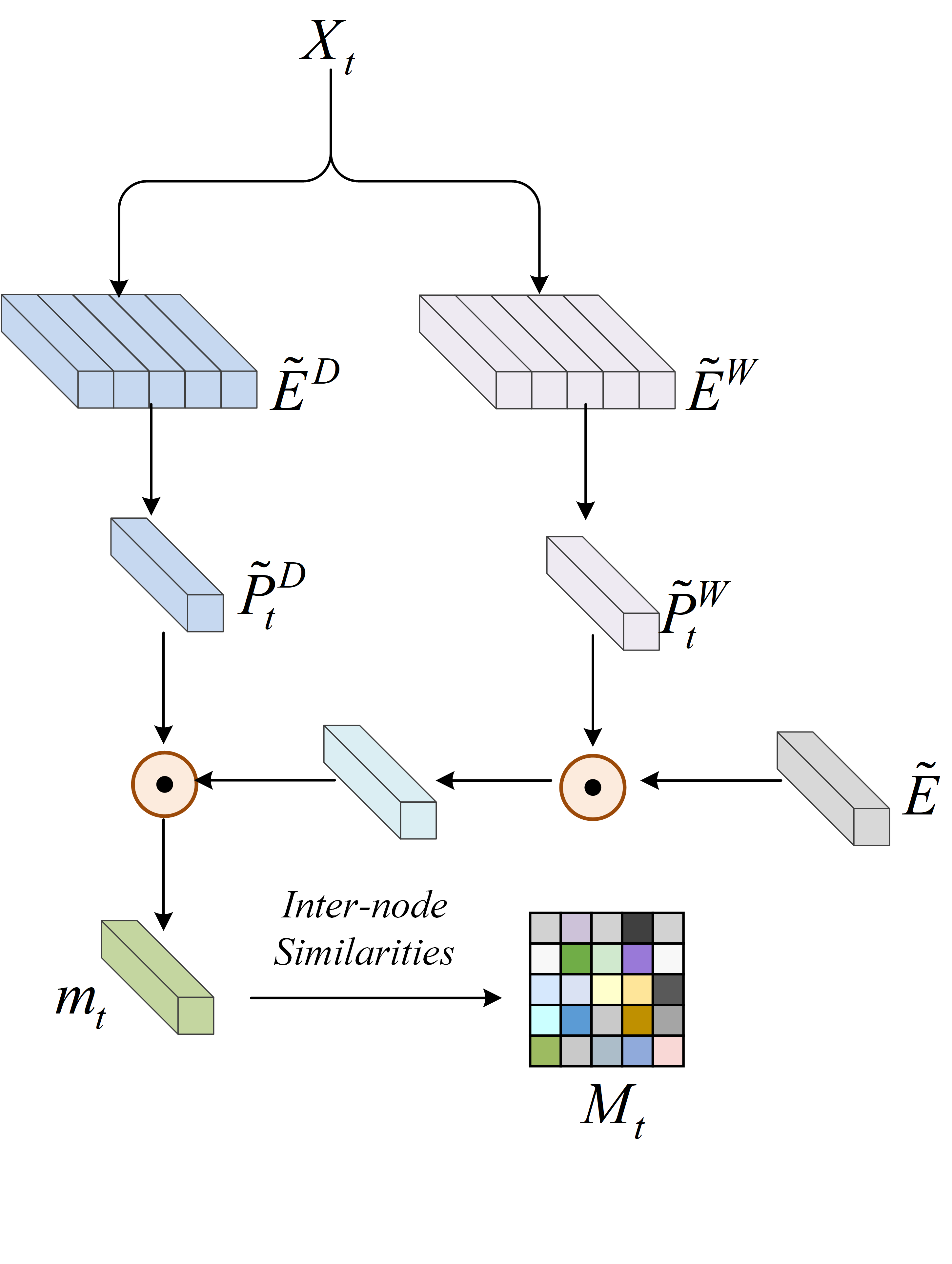}
         \caption{The process of generating a time-varying mask matrix at time step $t$}
         \label{fig:y equals x}
     \end{subfigure}
     \hfill
     \begin{subfigure}[b]{0.57\textwidth}
         \centering
         \includegraphics[width=\textwidth]{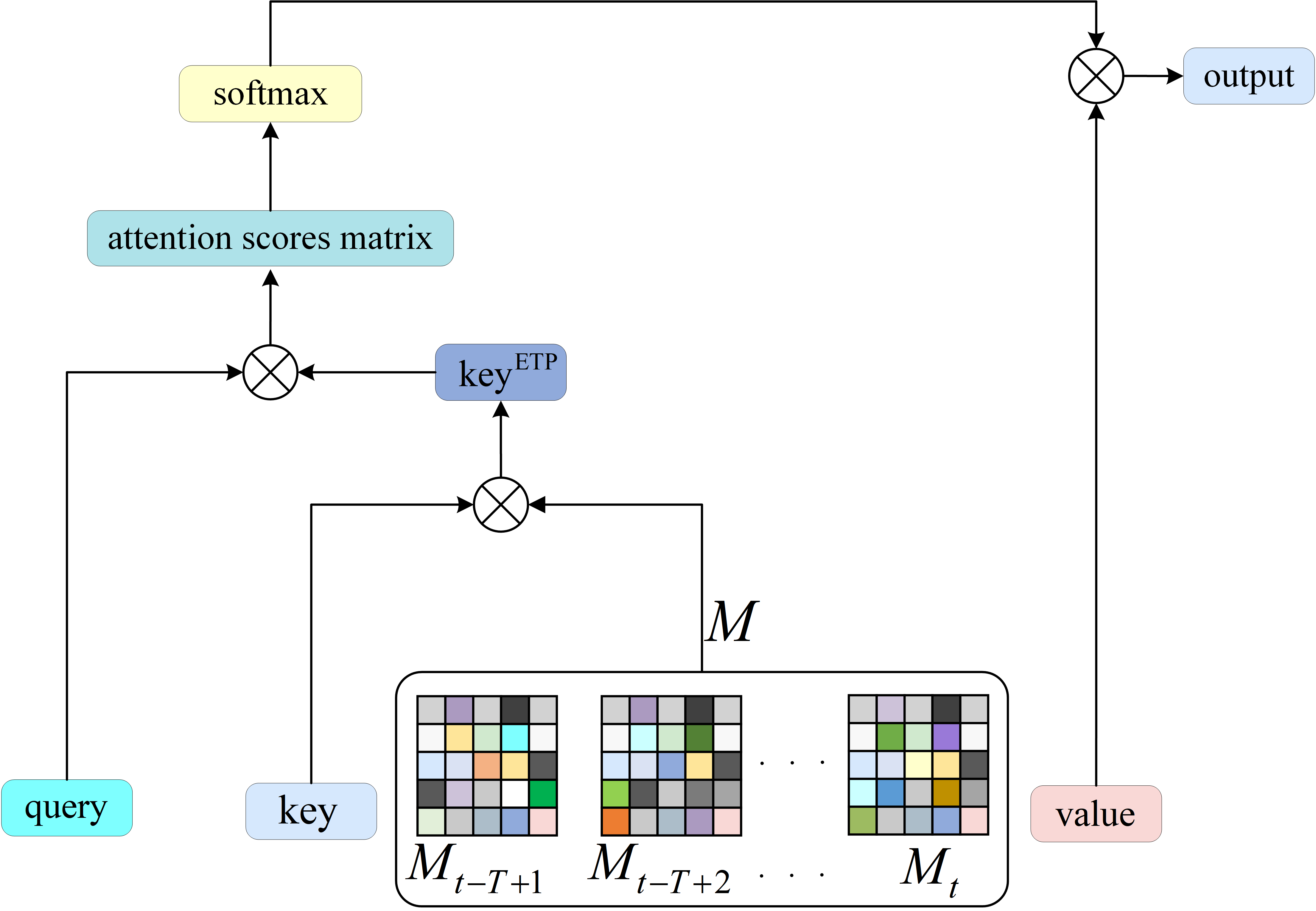}
         \caption{Enhanced temporal perception multi-head self-attention}
         \label{fig:y equals x}
     \end{subfigure}
     \hfill
     \caption{An illustration of the proposed enhanced temporal perception multi-head self-attention. \ (a) \ The process of generating a time-varying mask matrix at time step $t$, and the generation of similar time-varying mask matrices at other time steps are analogous to that of the time step $t$. \ (b) \ The key masking process and the enhanced temporal perception multi-head self-attention calculation process.}
        \label{fg:2}
\end{figure*}
\section{Problem Definition}
%
%
$\mathbf{Definition~1. ~Traffic
 \ Network.}$ The traffic network can be represented as a weighted directed graph $\mathcal{G}\ =\ (\mathcal{V},\mathcal{E},A)$, where $\mathcal{V}$ represents all nodes in the road network ($| \mathcal{V} | = N$). $\mathcal{E}$ is the set of all edges, which represents the connectivity between different nodes. $A\in R^{N \times N}$ represents the weighted adjacency matrix, and its value represents the degree of the correlation between different nodes. 

$\mathbf{Definition~2. ~Traffic \ Flow \ Characteristics.}$ Based on the traffic network,
the input of the traffic flow prediction is defined with $X \in R^{T \times N  \times C}$, which represents the traffic flow characteristics (such as traffic flow, speed, etc.) observed by all nodes at  $T$ times in the past. 
%

$\mathbf{Definition~3. ~Traffic \ Flow \ Prediction.}$ Given the road topology  $\mathcal{G}$  and traffic flow characteristics $ X=[X_{t-T+1},...,X_t] \in R^{T \times N  \times C}$ at  $T$ times in the past, our goal is to predict the traffic conditions $P=[P_{t+1},...,P_{t+\tau}] \in R^{\tau \times N  \times C}$ at $\tau$ times in the future by analyzing the traffic flow characteristics $X$.

$\mathbf{Definition~4. ~Time \-\ Varying \ Mask \ Embeding.}$ In the traffic network with $N$ nodes, where sensors sample data $n$ times a day ($n$ = 288 if the sampling interval is set to $5$ minutes) and considering a week of $7$ days. The dimension of the embedding is $d_m$. Define three independently trainable embeddings $\tilde E \in R^{N \times d_m}$, $\tilde E^D \in R^{n \times d_m}$ and $\tilde E^W \in R^{7 \times d_m}$. $\tilde E^D$ and $\tilde E^W$ are used to find and generate the daily sub-mask embedding $\tilde P_t^D \in R^{N \times d_m}$ and the weekly sub-mask embedding $\tilde P^W_t \in R^{N \times d_m}$ corresponding to the current traffic flow $X_t$. Meanwhile, $\tilde E$ is designed to optimize the generated daily sub-mask embedding $\tilde E^D$  and weekly sub-mask embedding $\tilde E^W$.
%
%
%
%
%
\section{Methodology}
%
%
In order to more accurately capture the dynamic spatio-temporal correlations between different traffic nodes, we have proposed a novel hybrid time-varying graph neural network (HTVGNN) for traffic flow prediction. The fundamental structure of HTVGNN is illustrated in Fig. \ref{fg:0001}. Firstly, an enhanced temporal perception multi-head self-attention mechanism is proposed to model dynamic temporal correlations. Then, a coupled time-varying graph convolution gated recurrent module (CTVGCRM) is employed to extract spatio-temporal correlations. In the CTVGCRM, we construct static and dynamic time-varying graphs to capture spatial correlations among traffic nodes. 

It worth noting that, more accurate temporal multi-attention can provide more differentiated spatio-temporal features for the coupled time-varying graphs. It helps the coupled time-varying graphs to learn traffic patterns at different time steps. In this way, our constructed coupled time-varying graphs can play a better effect. 
Detailed model structures and mechanisms of our proposed HTVGNN are summarized as follows. 
%
%
%
%
%

\subsection{An enhanced temporal perception multi-head self-attention}
%
%
%
%
Self-attention mechanism is a broader attention mechanism where query, key and value are different representations of the same sequence. In order to learn the complex and time-varying temporal dependencies of input data, the temporal feature $X \in R^{T \times N\times C}$ is usually mapped to a high-dimensional space with a widely used multi-head self-attention (MHSA).
For the $i^{th}$ head self-attention, $Q_i\in\ R^{T \times N \times d_q\ }, K_i\in\ R^{T \times N \times d_k\ }$ and $V_i\in\ R^{T \times N \times d_v \ }$ are generated by the linear projection as defined in Eq.(\ref{eq005}), respectively.
%
%
\begin{equation}\label{eq005}
\begin{split}
    & Q_i = X W_i^q,\\
    & K_i = X W_i^k, \\
    &V_i = X W_i^v,
\end{split}
\end{equation}
%
%
where $W_i^q$, $W_i^k$, and $W_i^v$ are learnable parameters of the $i^{th}$ self-attention head.

Then, $Q_i\in\ R^{T \times N \times d_q\ }, K_i\in\ R^{T \times N \times d_k\ }$ and $V_i\in\ R^{N \times d_v \ }$ are input into a parallel attention score function as defined in Eq.(\ref{eq006}) to calculate the attention of the $i^{th}$ self-attention head $head_i$  .
%
%
\begin{equation}\label{eq006}
    \begin{split}
    head_i &= Attention \left(Q_i,K_i,V_i \right) \\
     &= Softmax \left (\frac{Q_{i}  K_{i} }{\sqrt{d_{head}}} \right )V_{i},
     \end{split}
\end{equation}
%
%

Finally, the final multi-head self-attention corresponding to the input $X \in R ^{T \times N\times C}$ can be obtained by weighting the attention of the $h$ self-attention heads together. Specifically, it can be calculated by the following equation (\ref{eq007}),
%
%
\begin{equation}\label{eq007}
    MHSA=[head_1,\cdots,head_i,\cdots,head_h]W,
\end{equation}
%
%
where $h$ is the total amount of attention heads. $W$ is the final output projection matrix.
%
%
However, in traditional multi-head attention mechanism, the key ($K_{i}$) will cause unnecessary interference \cite{ref-4}. In addition, since the learned parameters $W_i^q$, $W_i^k$, and $W_i^v$ remained unchanged after training,  
%
%
the score calculated by the traditional multi-head attention equation defined in Eq. (\ref{eq006}) can not fully perceive the temporal trend of self-attention. Therefore, a time-varying mask embedding generator  is designed. The key process of the generator finds the daily  submask embedding $\tilde P^D$ and the weekly submask embedding $\tilde P^W$ of
the time corresponding to the current traffic flow {$X_t$}. Then, based on an independent trainable embedding $\tilde E \in R^{N \times d_m}$, we have proposed an enhanced temporal perception multi-head self-attention. For notational simplicity, our proposed enhanced temporal perception multi-head self-attention (ETPMSA), as shown in Fig.\ref{fg:2}, can be calculated as follows in Eq.(\ref{eq008}):
%
%
\begin{equation}
\begin{split}
    & {ETPMulAtt}(Q,K,V) =\Theta \cdot \widehat W,\\
    & \Theta= [head_1^{ETP},\cdots,head_i^{ETP},\cdots,head_h^{ETP}], \\
\end{split}
\label{eq008}
\end{equation}
%
%
%
where $\widehat W$ is a learnable parameter, $head_{i}^{ETP}$ denotes the $i^{th}$ self-attention head which can be calculted by the following Eq.(\ref{eq009}) to Eq.(\ref{eq013}).
%
%
\begin{equation}\label{eq009}
    head_i^{ETP} = Softmax \left (\frac{Q_i \cdot K_i^{ETP} }{\sqrt{d_{head}}} \right )V_i
\end{equation}
%
%
\begin{equation}\label{eq010}
K_i^{ETP}=M \odot K_{i},
\end{equation}
%
%
\begin{equation}\label{eq011}
    M =[M_{t-T+1},...,M_t]
\end{equation}
\begin{equation}\label{eq012}
M_t=m_t \odot m_t^T,
\end{equation}
%
%
\begin{equation}\label{eq013}
m_t = \tilde E \odot \Tilde P^D_t \odot \Tilde P^W_t,
\end{equation}
%
%
%
%
%
\begin{figure*}
     \centering
         \includegraphics[width=\textwidth]{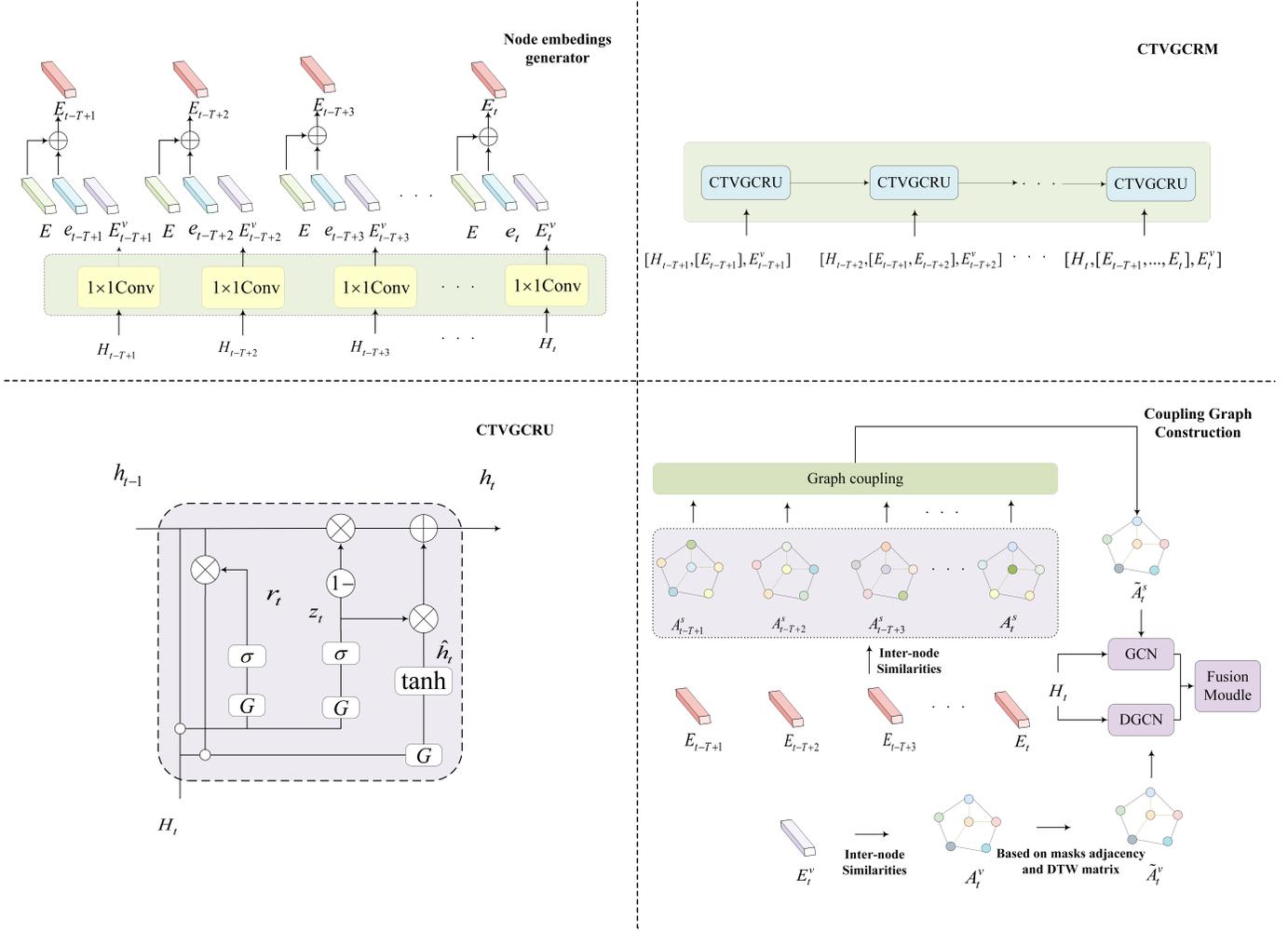}
         \caption{The components of CTVGCRM. In the figure $H_t$ is the output of the enhanced temporal perception multi-head self-attention at the time step $t$}
         \label{fg:3}
\end{figure*}
%
%
%
where $Q_i\in\ R^{T \times N \times d_q\ }, K_i\in\ R^{T \times N \times d_k\ }$ and $V_i\in\ R^{T \times N \times d_v \ }$ are the query, key and value of the traditional time multi-head attention as defined in equation (\ref{eq006}), $M_t \in R^{N \times N}$ is  mask matrix which is defined according to the size of the input data at the time step $t$.
It should be noted that, $m_t \in R^{N \times d_m}$ is the mask embedding which composed of three submask embedding $\tilde E \in R^{N \times d_m}$ and $\tilde P^D_t \in R^{N \times d_m}$ , $\tilde P^W_t \in R^{N \times d_m}$. 
Submask embedding $\Tilde P^{D}_t$ and $\Tilde P^{W}_t$ are the daily submask embedding $\Tilde P^{D}$ and weekly submask embedding $\Tilde P^{W}$ at $t$. The submask embedding $\Tilde P^{D}$ and $\Tilde P^{W}$ are employed to enhance the temporal perception ability of the model by updating the key $(K)$ of multi-heads self-attention mechanism.
%
%
\subsection{Position Encodings}
Temporal position coding is employed to represent the temporal order within a time series, where sequences of consecutive time steps exhibit higher correlation. In the conventional Transformer model, sine and cosine functions are utilized for mapping different feature dimensions during the calculation process of temporal position encoding, as formulated in Eq.(\ref{eq:0016}).
\begin{equation}\label{eq:0016}
    \begin{split}
& PE \left(pos,2d \right)\ =\ sin \left ( \frac{pos}{10000^\frac{2d}{D}\ } \right ) \\
& PE \left(pos,2d+1 \right)\ =\ cos \left (\frac{pos}{10000^{\frac{2d}{D}}\ } \right )\\
    \end{split}
\end{equation}

Among them, $pos$ represents the position and $d$ represents the feature dimension, $D$ denotes the embedding dimension of the model.
%
%
\subsection{Coupling time-varying graph convolution recurrent module}
%
%
In order to further extract the temporal and spatial characteristics, a novel coupling time-varying graph convolution recurrent module (CTVGCRM) is proposed.
The architecture of our proposed CTVGCRM is illustrated in Fig.~\ref{fg:3}. The details of the CTVGCRM are summarized as follows.
\subsubsection{Graph learning}
%
%
As pointed out in \cite{ref-3-1-1} \cite{ref-3-1-2}, there usually exist static and dynamic spatial correlations between different traffic nodes. In order to capture the static and dynamic spatial correlations, we have proposed a graph learning method to generate static graph and dynamic graph in a data-driven manner without providing any prior knowledge.
\begin{itemize}
  \item Static graph learning.
\end{itemize}

In order to capture the static spatial correlations between traffic nodes, we use two learnable embedding vectors to adaptively learn the static adjacency matrix at each time step. The static adjacency matrix $A_t^s$ of the $t^{th}$ time step is defined as Eq.(\ref{eq0012}):
%
%
\begin{equation}\label{eq0012}
\begin{split}
       & E_t=\ E\ +\ e_t \\
       &A_t^s=\ Softmax(E_t \odot E_t^T)\\
\end{split}
\end{equation}
%
%
where $E$ is a trainable node embedding, $\ e_t$ is a trainable bias of the node embedding $E$. Herein, in order to ensure the diversity of $A_t^s$ at different time step, we set different bias $e_t$ for different time step $t^{th}$ during the training process, $Softmax()$ is used as the activation function.

%
%
In order to model the similarity of traffic flow between different traffic nodes, a traffic pattern graph is defined by calculating the correlations of historical traffic flows using the DTW algorithm \cite{ref-3-2-1}. In general, traffic pattern graph reflects a long-term traffic pattern. However, short-term traffic pattern may play an more important role in short-term traffic flow prediction. Therefore, in order to obtain more spatial topology-related information at different time steps, we have proposed a static graph learning module to improve the learning ability of the static adjacency matrix. At each time step, we can construct a static graph with the static adjacency matrix $A_t^s$ defined in Eq.(\ref{eq0012}). Then, in order to more accurately model static spatial correlations between different traffic nodes, we can further calculate the coupling time-varing adjacency matrix by the following Eq.(\ref{eq0013}):
%
%
\begin{equation}\label{eq0013}
    \begin{split}
        &\tilde A_{t-T+1}^s= A_{t-T+1}^s = Softmax((E_{t-T+1} \\
        &\odot E_{t-T+1}^T)\\
        &\tilde A_{t-T+2}^s=[A_{t-T+1}^s ||Softmax(E_{t-T+2}\\
        &\odot E_{t-T+2}^T)] \odot W_{t-T+2}\\
        &  ...... \\
        &\tilde A_{t}^s=[A_{t-T+1}^s ||A_{t-T+2}^s ||...||A_{t-1}^s ||Softmax(\\
        &E_{t} \odot E_{t}^T)] \odot W_{t}
    \end{split}
\end{equation}
%
%
%
where $W_{t-T+2},....W_{t}$ are learnable parameter.after completing the graph coupling process,The coupling graph  $\tilde A_t^s$ is utilized as the ultimate static adjacency matrix for spatial correlation modeling at each time step.
%
%
\begin{itemize}
  \item Dynamic graph learning.
\end{itemize}

In order to capture the dynamic spatial correlations between different traffic nodes, the dynamic adjacency matrix is calculated based on the hidden embedding $H$ of current traffic features. Herein, we design a 1$\times$1 convolution layer as the mapping function $\phi$. The mapping processing can be defined by the Eq.(\ref{eq014}). Then, a learnable vector $a$ is defined to calculate the attention weight between node $i$ and node $j$.

the dynamic adjacency matrix $\tilde A^v_{t_{(i,j)}}$ at the $t^{th}$ time step between node $i$ and  node $j$ is defined as Eq.(\ref{eq014}) to Eq.(\ref{eq0017})
%
%

%
%
\begin{equation}\label{eq014}
   E^v_t \ =\ H_t W_{\phi}
\end{equation}
%
%
\begin{equation}\label{eq015}
   e^v_{ij}\ \ = a\cdot (\left[E^v_t\right]_i\ ||\left[E^v_t\right]_j)
\end{equation}
%
%
where $H_t$ is the value of the hidden embedding $H \in R^{T \times N \times C_{\phi}}$ at time step $t$. $E^v_t \in R^{N \times d_{\phi}}$ is the mapped traffic features including short-term dynamic spatial traffic patterns implicit in current traffic features $X$. $\left[E^v_t\right]_i$ is the $i^{th}$ node in $E^v_t$. $||$ represents a splicing operation. $e^v_{ij}$\ \ indicates the dynamic traffic similarity between node $i$ and node $j$ in the road network.Then, using $Softmax()$ for normalization, the generation process of $A^v_t(i,j)$ is as Eq.(\ref{eq016}):
%
%
\begin{equation}\label{eq016}
      A^v_t(i,j)\ \ = \left[Softmax(e^v_{i,:}) \right]_j
\end{equation}
%
%
where $A^v_t(i,j)$ is the value of $A^v_t$ at node $i$ and node $j$, $A^v_t$ is the value of $A^v$ at time step $t$.

Then, Dynamic changes of spatial correlations occuring in the current time should belong to the actual existence and most active nodes. such as nodes which located in adjacency matrix $A$ and DTW matrix $A_{dtw}$. Thus, we combine the adjacency matrix $A$ and DTW matrix $A_{dtw}$ as the mask matrix of dynamic spatial correlations. Finally the dynamic adjacency matrix $\tilde A^v_t$ can be define as Eq.(\ref{eq0017}):
%
%
\begin{equation}\label{eq0017}
     \tilde A^v_t\ \ = A^v_t \odot \left ( A + A_{dtw} \right)
\end{equation}
where $A^v_t$ is the value of $A^v$ at time step $t$,$\tilde A^v_t$ is the value of $\tilde A^v$ at time step $t$.
%
%

%
%
%
%
\subsubsection{Coupling time-varying graph convolution gated recurrent unit}
%
%
GRU is a variant of RNN which maintains the effectiveness of
RNN in capturing temporal features while simplifying the
parameters of the model, making it widely used in capturing
temporal correlation. Based on the previous studies \cite{ref10} \cite{ref17}, a coupling time-varying graph convolution gated recurrent
unit (CTVGCRU) is constructed by replacing the matrix multiplication in GRU with a combination of graph convolutional network and an node adaptive parameter learning module. The specific form can
be defined as Eq.(\ref{eq0018}):
%
%
\begin{equation}\label{eq0018}
\begin{split}
   & \widehat H_t = H_t||h_{t-1}\\
   &\widehat E = [E_{ t-T+1 },...,E_t]\\
   & {z}_t=\sigma(||\left[\mathcal{G}_s([\widehat H_t,\widehat E];\theta_{z_1}),\mathcal{G}_d ([\widehat H_t,E_t^v];\theta_{z_2}) \right])\\
& r_t=  \sigma(||\left[\mathcal{G}_s([\widehat H_t,\widehat E];\theta_{r_1}),\mathcal{G}_d ([\widehat H_t,E_t^v];\theta_{r_2}) \right])\\
& c_t=\tanh(||\left[\mathcal{G}_s([ g_t,\widehat E];\theta_{c_1}),\mathcal{G}_d ([g_t,E_t^v];\theta_{c_2}) \right])\\
& g_t = H_t||{r}_t \odot h_{t-1}\\
& h_t = z_t \odot\ h_{t-1} + (1 - z_t) \odot c_t
\end{split}
\end{equation}
%
%
where $H_t$ and $h_t$ are the input hidden embedding and output hidden embedding at the time step $t$, respectively. $||$ represents the concatenation operation, ${z}_t$ and ${r}_t$ represent reset gate and update gate the time step $t$, respectively. \ $\mathcal{G}_s$ and $\mathcal{G}_d$ represent the graph convolution network (GCN) and dynamic graph convolution network (DGCN), respectively. $\theta_{z_1}$, $\theta_{z_2}$ and $\theta_{r_1}$, $\theta_{r_2}$ and \ $\theta_{c_1}$, $\theta_{c_2}$ are learnable parameters in CTVGCRM.
\begin{table*}[htbp]
\centering
\caption{The details of  the datasets}
\begin{tabular}{ccccccccccccccccccccc}     \hline
\multicolumn{4}{c}{\multirow{1}*{Dataset}}     & \multicolumn{4}{c}{\multirow{1}*{PEMS03}}  & \multicolumn{4}{c}{\multirow{1}*{PEMS04}}  & \multicolumn{4}{c}{\multirow{1}*{PEMS07}}& \multicolumn{4}{c}{\multirow{1}*{PEMS08}}\\ \hline
  \multicolumn{4}{c}{\# Nodes}                             &  \multicolumn{4}{c}{358}    &    \multicolumn{4}{c}{307} &  \multicolumn{4}{c}{883}    &    \multicolumn{4}{c}{170}  \\
  \multicolumn{4}{c}{\# Edges}                             &  \multicolumn{4}{c}{866}    &    \multicolumn{4}{c}{340} &  \multicolumn{4}{c}{340}    &    \multicolumn{4}{c}{277}  \\
            \multicolumn{4}{c}{Time steps}                             &  \multicolumn{4}{c}{ 26208}   &  \multicolumn{4}{c}{16992}  &  \multicolumn{4}{c}{28224}   &  \multicolumn{4}{c}{17856}     \\
      \multicolumn{4}{c}{Time span}                             &  \multicolumn{4}{c}{ 2018/9/1–2018/11/30}   &  \multicolumn{4}{c}{ 2018/1/1–2018/2/28}  &  \multicolumn{4}{c}{ 2017/5/1–2017/8/31}   &  \multicolumn{4}{c}{ 2016/7/1–2016/8/31}     \\

        \multicolumn{4}{c}{Missing ratio}                             &  \multicolumn{4}{c}{ 0.672\%}   &  \multicolumn{4}{c}{3.182\%}  &  \multicolumn{4}{c}{0.452\%}   &  \multicolumn{4}{c}{0.696\%}  \\
   \multicolumn{4}{c}{ Time interval}                             &  \multicolumn{16}{c}{5 min}\\
   \multicolumn{4}{c}{daily range}                             &  \multicolumn{16}{c}{0:00-24:00}\\
       \hline

\end{tabular}\label{tb:1}
\end{table*}
%
%
%
%
\begin{table*}[htbp]
\centering
\renewcommand\arraystretch{1.1}
\tabcolsep=0.02cm
\caption{Evaluation results of each model on the datasets PEMS03 and PEMS04 under different prediction steps }
\begin{tabular}{cccccccccccccccc}     \hline
\multirow{2}*{Data}   & \multicolumn{3}{c}{\multirow{2}*{Model}}  & \multicolumn{3}{c}{15min} & \multicolumn{3}{c}{30min} & \multicolumn{3}{c}{60min}  & \multicolumn{3}{c}{Average} \\ \cline{5-16}
                        & \multicolumn{3}{c}{}                       & MAE    &MAPE(\%)   & RMSE        & MAE     &MAPE(\%)    & RMSE        & MAE     &MAPE(\%)    & RMSE        & MAE      &MAPE(\%)   & RMSE                \\ \hline
\multirow{11}*{PEMS03} & \multicolumn{3}{c}{HA \cite{ref19}}                     &     23.57  &    22.13 &36.48       &  27.51  &     29.14    & 44.99         &    40.11  &    39.10  & 61.16        & 30.03     &   28.63    & 47.17      \\
                        & \multicolumn{3}{c}{VAR \cite{ref-14}}                   &    17.41  &  18.20   & 25.42        & 22.13     & 24.28    & 32.20        &  31.65  & 37.42& 44.89        &     22.91  & 25.53    & 33.04          \\
                        & \multicolumn{3}{c}{LSTM \cite{ref--6}}
                        & 16.69  &  16.02 & 25.54 & 20.03  & 19.34 & 30.40 & 27.42  & 29.48 & 40.20 & 20.64  & 20.57 & 31.63 \\
                        & \multicolumn{3}{c}{DCRNN \cite{ref21}}
                         &15.45  & 14.98 &25.75&17.98 & 17.38  &29.80&23.12  & 22.54  & 37.18&18.31 &17.74&30.08   \\
                        & \multicolumn{3}{c}{STGCN \cite{ref-3-2}}
                        & 18.46 & 19.56&31.27& 19.55 &22.74& 32.59& 21.70  &24.08 &35.29&19.64  &21.63&32.74   \\
                        & \multicolumn{3}{c}{GWN \cite{ref9}}
                        & 16.06  & 15.39 & 27.38& 19.25  & 18.45 & 32.29& 26.12  & 26.32 &  41.89& 19.82  & 19.21 & 32.88\\
                        & \multicolumn{3}{c}{ASTGCN \cite{ref24}}
                        & 15.24  &15.31 &25.25&14.63 &16.71  &15.65&27.64  &20.25 &18.63&16.96  &15.97 &28.15  \\
                        & \multicolumn{3}{c}{AGCRN \cite{ref10}}               &   14.61  &  13.82 & 25.84       &   15.65   &  14.61   & 27.51       &   17.42   & 15.94    & 30.23       &  15.70    &  14.66 & 27.63   \\
                           & \multicolumn{3}{c}{GMAN \cite{ref26}}
                         &15.65   & 16.68 & 26.09& 16.32  & 16.32 & 27.15& 17.73  & 18.91 & 29.26& 16.45  & 17.47 & 27.31\\
                           & \multicolumn{3}{c}{Tformer \cite{ref27}}               &  14.97    &  15.63   & 23.86         &  16.54 & 17.15     & 26.22         & 19.08   & 19.49     & 29.57       &
 16.54 & 17.15    & 26.25   \\
  & \multicolumn{3}{c}{DDSTGCN \cite{ref28}}              &13.70  & 13.99 & 23.95 & 14.88  & 14.91 & 25.38  & 16.88 & 16.63 & 29.13  & 14.93 & 14.96 & 25.83\\

     & \multicolumn{3}{c}{Bi-STAT \cite{ref30}}              & 14.30  & 14.78 & 25.07 & 15.43  & 15.67 & 26.93  & 17.19 & 17.21 & 29.57  & 15.47 & 15.74 & 26.92\\

       & \multicolumn{3}{c}{STWave \cite{ref-30}}              &13.91  & 14.90 & 24.82 & 14.92  & 15.53 & 26.70  & 16.68 & 18.92 & 29.19  & 14.96 & 16.54 & 26.62\\
  & \multicolumn{3}{c}{HTVGNN}              & $\mathbf{13.09}$  &  $\mathbf{13.65}$    &  $\mathbf{22.43}$        &  $\mathbf{14.30}$  &  $\mathbf{14.69}$     & $\mathbf{24.59}$       & $\mathbf{15.97}$  &  $\mathbf{16.17}$    & $\mathbf{27.04}$       & $\mathbf{14.20}$   & $\mathbf{14.52}$ & $\mathbf{24.26}$ \\ \hline

\multirow{11}*{PEMS04} & \multicolumn{3}{c}{HA \cite{ref19}}                     &     30.70  &    20.71 & 45.02       &  37.47  &     27.10    & 54.37         &    50.84   &    38.58  & 72.67        & 38.56     &   28.17    & 56.85       \\
                        & \multicolumn{3}{c}{VAR \cite{ref-14}}
                         &22.75  & 16.95 & 33.11& 27.93   & 22.10 & 39.68& 38.26   &32.74 & 52.57& 28.78  & 23.01 & 40.72         \\
                        & \multicolumn{3}{c}{LSTM \cite{ref--6}}
                        & 21.89  & 14.79 & 33.01& 25.98  &18.07 & 38.65& 35.12  & 27.28 & 50.35& 26.81 & 19.28 & 40.18\\
                        & \multicolumn{3}{c}{DCRNN \cite{ref21}}
                        &20.61  &13.91 &32.22&23.87  &16.18 &36.80&30.83  &21.27 &46.27&24.44  &16.64&37.48\\
                        & \multicolumn{3}{c}{STGCN \cite{ref-3-2}}
                        &23.43  &20.43 &35.30&25.45  &22.56 &37.83&30.45  &27.84 &44.12&25.91  &22.90&38.44\\
                        & \multicolumn{3}{c}{GWN \cite{ref9}}
                         &20.90  &14.44 &33.04&24.53  &17.45 &38.22&32.58  &23.38 &49.15&25.25  &18.22 &39.10\\
                        & \multicolumn{3}{c}{ASTGCN \cite{ref24}}
                        &19.55  &12.99 &31.16&20.32  &13.39 &32.39&23.63  &15.31 &37.21&20.59  &13.53&32.85\\
                        & \multicolumn{3}{c}{AGCRN \cite{ref10}}               &  18.88   &  12.49 & 30.92       &   19.65   &  12.94   & 32.29       &   21.34   & 13.87    & 34.98       &  19.76    &  12.99 & 32.48   \\

                            & \multicolumn{3}{c}{GMAN \cite{ref26}}
                         & 18.54  &12.99 &29.74&19.09  &13.47 &30.70&20.92  &15.12 &33.24 &19.34  &13.71&30.98\\
                           & \multicolumn{3}{c}{Tformer \cite{ref27}}               &  21.02    & 14.94  & 31.33         &  22.37 & 16.24     & 33.05        & 23.72   &  17.41     &  34.73       &
 21.10 &  15.13    & 31.46  \\
 & \multicolumn{3}{c}{DDSTGCN \cite{ref28}}               &  18.40    & 12.86   & 29.38         &  19.64 & 13.77     & 31.15        & 21.82  & 15.40    & 34.07      &
 19.74 & 13.83    & 31.12   \\

 & \multicolumn{3}{c}{Bi-STAT \cite{ref30}}               &  18.16    & 12.39   & 29.29         &  19.01 & 12.92     & 30.61        & 21.68  & 15.36    & 33.85      &
 19.66 & 13.02    & 30.64   \\
   & \multicolumn{3}{c}{STWave \cite{ref-30}}               &  17.64    & 11.89   & 28.98         &  18.68 & 12.62   & 30.62        & 20.03  &  13.68    & 32.64    & 18.64 & 12.62    & 30.55   \\
  & \multicolumn{3}{c}{HTVGNN}              & $\mathbf{17.25}$ & $\mathbf{11.44}$     & $\mathbf{28.48}$        &   $\mathbf{18.01}$  &  $\mathbf{11.89}$     & $\mathbf{29.81}$      & $\mathbf{19.20}$   &  $\mathbf{12.72}$    & $\mathbf{31.62}$        & $\mathbf{17.99}$   & $\mathbf{11.90}$  & $\mathbf{29.74}$ \\ \hline

\end{tabular}\label{tb:2}
\end{table*}

\begin{table*}[htbp]
\centering
\renewcommand\arraystretch{1.1}
\tabcolsep=0.02cm
\caption{Evaluation results of each model on the datasets PEMS07 and PEMS08 under different prediction steps }
\begin{tabular}{cccccccccccccccc}     \hline
\multirow{2}*{Data}   & \multicolumn{3}{c}{\multirow{2}*{Model}}  & \multicolumn{3}{c}{15min} & \multicolumn{3}{c}{30min} & \multicolumn{3}{c}{60min}  & \multicolumn{3}{c}{Average} \\ \cline{5-16}
                        & \multicolumn{3}{c}{}                       & MAE    &MAPE(\%)   & RMSE        & MAE     &MAPE(\%)    & RMSE        & MAE     &MAPE(\%)    & RMSE        & MAE      &MAPE(\%)   & RMSE                \\ \hline

\multirow{11}*{PEMS07} & \multicolumn{3}{c}{HA \cite{ref19}}                     &     35.67  &    16.49 & 52.07       &  44.06  &    20.82    & 64.02         &    60.44   &    29.69  & 87.05       & 45.36     &   21.59    & 67.17       \\
                        & \multicolumn{3}{c}{VAR \cite{ref-14}}
                        & 25.10  &10.86 &37.22&31.30  &14.08 &45.23&42.96  &42.96 &60.04&32.03  &14.61 &46.12\\
                        & \multicolumn{3}{c}{LSTM \cite{ref--6}}
                         &23.66  &10.48 &35.77&28.50  &12.76 &42.67 &39.12  &18.66 &56.60&29.32  &13.30&44.39\\
                        & \multicolumn{3}{c}{DCRNN \cite{ref21}}
                        &21.97  &9.42 &33.94&25.68  &11.14 &39.16&33.29  &14.93 &49.09&26.15  &11.44&39.57\\
                        & \multicolumn{3}{c}{STGCN \cite{ref-3-2}}
                         &32.52  & 16.23 &51.71&33.26  &16.25 &52.75&36.02  &17.59 &55.64& 33.62  &16.53 &53.04\\
                        & \multicolumn{3}{c}{GWN \cite{ref9}}
                         &21.72  &9.52 &34.78&25.89  &11.89 &41.05&34.83  &16.78 &53.72&26.58  &12.28 &41.85\\
                        & \multicolumn{3}{c}{ASTGCN \cite{ref24}}
                         &23.94  &10.25 &36.88&28.77  &12.50 &43.89&39.37  &18.00 &58.07&29.59  &13.04 &45.52\\
                        & \multicolumn{3}{c}{AGCRN \cite{ref10}}               &  19.89   &  8.34 & 32.68       &   21.29   &  8.93   & 34.91       &   23.51   & 9.85    & 38.45       &  21.26   &  8.91 & 34.99   \\

                            & \multicolumn{3}{c}{GMAN \cite{ref26}}
                         & 19.08 & 8.11 & 31.23 & 20.23  & 8.57 & 33.37&22.30  &9.56 &36.58&20.35	 &8.65&33.39 \\
                           & \multicolumn{3}{c}{Tformer \cite{ref27}}               &  20.28    &  9.07  & 31.31        & 22.10 & 10.08     & 34.19       & 25.15   & 11.87    & 38.52      &
 22.07 & 10.12    & 34.21   \\
 & \multicolumn{3}{c}{DDSTGCN \cite{ref28}}               & 19.72   & 8.82   & 31.83         & 27.42  & 11.87   & 41.87     & 24.49  & 10.91     &  39.03       &
 21.55 & 9.59    & 34.65  \\

   & \multicolumn{3}{c}{Bi-STAT \cite{ref30}}
                            &19.66  &8.36 &31.81&21.44  &9.07 &34.70&24.34  &10.43 &38.88&21.50  &9.15&34.64\\
    & \multicolumn{3}{c}{STWave \cite{ref-30}}
                            &18.23  & 7.61  &30.95&19.48  &8.16 &33.32&21.45  &9.07 &36.55&19.60  &8.21&33.29\\
  & \multicolumn{3}{c}{HTVGNN}              &  $\mathbf{17.99}$ & $\mathbf{7.60}$      & $\mathbf{30.00}$ & $\mathbf{19.52}$     & $\mathbf{8.15}$ & $\mathbf{32.65}$      &    $\mathbf{21.86}$ & $\mathbf{9.21}$     &  $\mathbf{36.20}$ & $\mathbf{19.46}$      &  $\mathbf{8.15}$ & $\mathbf{32.53}$       \\\hline

\multirow{11}*{PEMS08} & \multicolumn{3}{c}{HA \cite{ref19}}                     &     25.15  &    15.76 & 37.00        &  31.09  &     19.62    & 45.32         &    42.92   &    27.66  & 61.42        & 32.06     &   20.35    & 47.52       \\
                        & \multicolumn{3}{c}{VAR \cite{ref-14}}
                         &17.88  &12.17 &26.48&22.21&15.50  &32.39 &30.63  &21.94 &42.92&22.81&15.96 &32.97 \\
                        & \multicolumn{3}{c}{LSTM \cite{ref--6}}
                        &17.71  &11.58 &26.65&21.31  &14.83 &31.97&29.37  &19.62 &42.53&21.96 &14.52 &33.22 \\
                        & \multicolumn{3}{c}{DCRNN \cite{ref21}}
                         &15.93  &10.05 &24.62&18.30  &11.57 &28.43&23.02  &14.42 &35.05&18.59  &11.72 &28.64   \\
                        & \multicolumn{3}{c}{STGCN \cite{ref-3-2}}
                        &20.53  &14.25 &29.92&22.21  &16.23 &32.24&26.40  &21.17 &37.37&22.62 &16.52 &32.66\\
                          & \multicolumn{3}{c}{GWN \cite{ref9}}
                          &15.59  &9.84 &25.27&18.28  &12.06 &30.02&24.18 & 15.63  &38.85&18.75  &12.12 &30.47\\
                        & \multicolumn{3}{c}{ASTGCN \cite{ref24}}
                        &16.36  &10.01 &25.24&18.38  & 11.12 &28.34&22.60  &13.57 &34.39&18.61  &11.27&28.82\\
                        & \multicolumn{3}{c}{AGCRN \cite{ref10}}               &  14.90   &  9.67 & 23.39       &   15.90   &  10.26   & 25.18       &   17.84   & 11.37    & 28.32       &  16.01   &  10.31 & 25.36   \\
                            & \multicolumn{3}{c}{GMAN \cite{ref26}}
                          &13.75  &9.07 &22.84&14.16	  &9.35 &23.74&15.47  &10.40 &25.72&14.32  &9.50&23.89\\
                           & \multicolumn{3}{c}{Tformer \cite{ref27}}               &   15.41   &  10.14   & 22.98       & 16.83 &  11.77     &  25.11       & 19.08   &  12.61   &   28.19        &
 16.79 &  11.41    &  25.11   \\
 & \multicolumn{3}{c}{DDSTGCN \cite{ref28}}               &  14.31    &  9.71   & 22.45         &  15.44 &  10.39     & 24.42        & 17.42  & 11.61     & 27.33       &
 15.49 & 10.42    & 24.36   \\

  & \multicolumn{3}{c}{Bi-STAT \cite{ref30}}               &  13.92    &  9.02   & 22.32         &  14.64 &  9.52     & 23.75        & 16.14  & 10.59     & 26.11       &
 14.76 & 9.62    & 23.80   \\

 & \multicolumn{3}{c}{STWave \cite{ref-30}}               &  12.77    &  8.50   & 21.71      &  13.69 &  9.40     & 23.47  & 14.98  & 10.44     & 25.85       &
 13.70 & 9.30    & 23.47   \\
 & \multicolumn{3}{c}{HTVGNN}              &  $\mathbf{12.41}$ & $\mathbf{8.09}$      & $\mathbf{21.06}$ & $\mathbf{13.28}$     & $\mathbf{8.65}$ & $\mathbf{22.83}$      &    $\mathbf{14.48}$ & $\mathbf{9.44}$     &  $\mathbf{24.96}$ & $\mathbf{13.24}$      &  $\mathbf{8.63}$ & $\mathbf{22.67}$ \\ \hline

\end{tabular}\label{tb:3}
\end{table*}
%
%
%
%
%
\section{Experiments}

\subsection{Datasets}
%
%

%
%
We validate the predictive performance of our proposed method HTVGNN by conducting experiments on four real-world traffic datasets, namely PEMS03, PEMS04, PEMS07 and PEMS08. These datasets were collected by the California Department of Motor Transportation Performance Measurement System (PeMS). Please refer to Table \ref{tb:1} for detailed information.

%
%

\subsection{Evaluation Metrics}

In this study, we employed three effectiveness measures, namely Mean Absolute Error (MAE), Mean Absolute Percentage Error (MAPE), and Root Mean Square Error (RMSE), to assess the accuracy of the HTVGNN model as described in Eq.(\ref{eq0020}) to (\ref{eq0022}).
%
%
\begin{equation}\label{eq0020}
    MAPE \ = \ \frac{1}{n}\ \sum_i \left|{\frac{y_i \ - \ \hat y_i}{{y_i}}}\right| \times 100\%,
\end{equation}
%
%
%
%
\begin{equation}\label{eq0021}
  MAE = \frac{1}{n} \sum_i |\hat y_i-y_i|,
\end{equation}
\begin{equation}\label{eq0022}
    RMSE =  \sqrt{\frac{1}{n} \sum_i  (\hat y_i - y_i)^2},
\end{equation}
%
%
where $\hat y_i$ and $y_i$ represent the predicted value and real value of traffic flow at node $i$, respectively. $n$ indicates the total number of nodes. 

\subsection{Baselines}
We compare the HTVGNN model against thirteen state-of-the-art baselines as outlined below:

$\bullet  \ \mathbf{Historical  \ Average \ (HA)}$ \cite{ref19}
The HA model takes the traffic condition
Modeled as a seasonal process and using historical averages as forecast.

$\bullet  \ \mathbf{Vector \ Auto \ Regression \ (VAR)}$ \cite{ref-14}VAR model can handle time series models of correlations between multiple variables.

$\bullet  \ \mathbf{Long \ Short \ Term \ Memory \ (LSTM)}$ \cite{ref--6} LSTM is a classical
variant of RNN, and it can be used for long-term time series forecasting.

$\bullet  \ \mathbf{Spatial- \ temporal \ graph \ convlutional \ networks} \\ \mathbf{(STGCN)}$ \cite{ref-3-2}STGCN is a novel spatio-temporal graph convolutional network that effectively captures both spatial and temporal features through the integration of graph convolution and gated time convolution.
$\bullet  \ \mathbf{Diffusion} \ \mathbf{Convolutional} \ \mathbf{Recurrent} \ \mathbf{Neural} \ \mathbf{Network} \\
\mathbf{(DCRNN)}$ \cite{ref21} DCRNN models the spatial dependency by
relating traffic flow to a diffusion process, and introduce diffusion convolution into GRU in an encoder-decoder manner
for multi-step traffic flow forecasting.

$\bullet  \ \mathbf{Graph \ WaveNet(GWN)}$ \cite{ref9} Graph WaveNet for Deep
Spatial-Temporal Graph Modeling uses adaptive adjacency matrices and learns by node embedding. Temporal dependence captures using 1D dilated convolution.

$\bullet  \ \mathbf{Attention \ Based \ Spatial-Temporal \ Graph \ Con} \\
\mathbf{volutional \ Network \ (ASTGCN)}$ \cite{ref24} ASTGCN introduces attention
mechanisms into CNN and GCN to respectively model spatial
 and temporal dynamics of trafﬁc data.

 $\bullet  \ \mathbf{Adaptive \ Graph \ Convolutional \ Recurrent \ Net-} \\ \mathbf{work}
 \mathbf{(AGCRN)}$ \cite{ref10} AGCRN utilizes a node adaptive parameter
 learning to enhance GCN and combines it with GRU to capture spatial and temporal features of trafﬁc data.


$\bullet  \ \mathbf{A \ Graph \ Multi-Attention \ Network \ for \ Traffic} \\
\mathbf{Prediction \ (GMAN)}$ \cite{ref26} GMAN uses spatial and temporal multi-attention mechanisms to model dynamic spatio-temporal correlations, and a gating mechanism is designed to better fuse spatio-temporal features.

     $\bullet  \ \mathbf{Learning} \ \mathbf{Dynamic} \ \mathbf{and} \ \mathbf{Hierarchical} \ \mathbf{Traffic} \ \mathbf{Spatio} \\
\mathbf{temporal} \ \mathbf{Features} \ \mathbf{With} \ \mathbf{Transformer} \ (\mathbf{Tformer})$ \cite{ref27} LSTM is used to encode historical stream data, and a global encoder and a global-local decoder are designed to extract global and local spatial correlations.

     $\bullet  \  \mathbf{Dual} \  \mathbf{Dynamic} \  \mathbf{Spatial-Temporal}   \  \mathbf{Graph} \  \mathbf{Convo-} \\ \mathbf{lution} \
     \mathbf{Network} \  \mathbf{for} \  \mathbf{Traffic} \  \mathbf{Prediction} \ (\mathbf{DDSTGCN})$ \cite{ref28} a novel dual dynamic GCN traffic prediction method. DDSTGCN transforms the edges of traffic flow graph to its dual hypergraph and exploits their spatial and temporal property by GCN on hypergraph.
          $\bullet  \ \mathbf{Bidirectional} \ \mathbf{Spatial-Temporal} \ \mathbf{Adaptive} \ \mathbf{Tr-} \\
          \mathbf{ansformer} \ \mathbf{{for}} \ \mathbf{Urban}  \ \mathbf{Traffic}   \ \mathbf{Flow}  \ \mathbf{Forecasting} \ (\mathbf{Bi-} \\ \mathbf{STAT})$ \cite{ref30} The encoder-decoder structure consists of a temporal Transformer and a spatial Transformer. The difference is that a recall module is added to the decoder to provide supplementary information for the prediction task. The designed DHM module is used to dynamically adjust the complexity of the model according to the complexity of the prediction task.
          $\bullet  \ \mathbf{Decoupled}  \ \mathbf{Traffic} \ \mathbf{Prediction} \ \mathbf{with}  \ \mathbf{Efficient}   \\ \mathbf{Spectrum-}  \mathbf{based} \ \mathbf{Attention} \ \mathbf{Networks} \ (\mathbf{STWave})$ \cite{ref-30} STWave decomposes complex traffic data into stable trends and fluctuating events, then models trends and events separately using dual-channel spatio-temporal networks. Finally, by fusing trends and events, we can predict reasonable future traffic flow. Additionally, we incorporate a novel query sampling strategy and graph wavelet-based graph position encoding into graph attention networks to efficiently and effectively model dynamic spatial correlations.

\begin{figure*}
     \centering
     \begin{subfigure}[b]{0.24\textwidth}
         \centering
         \includegraphics[width=\textwidth]{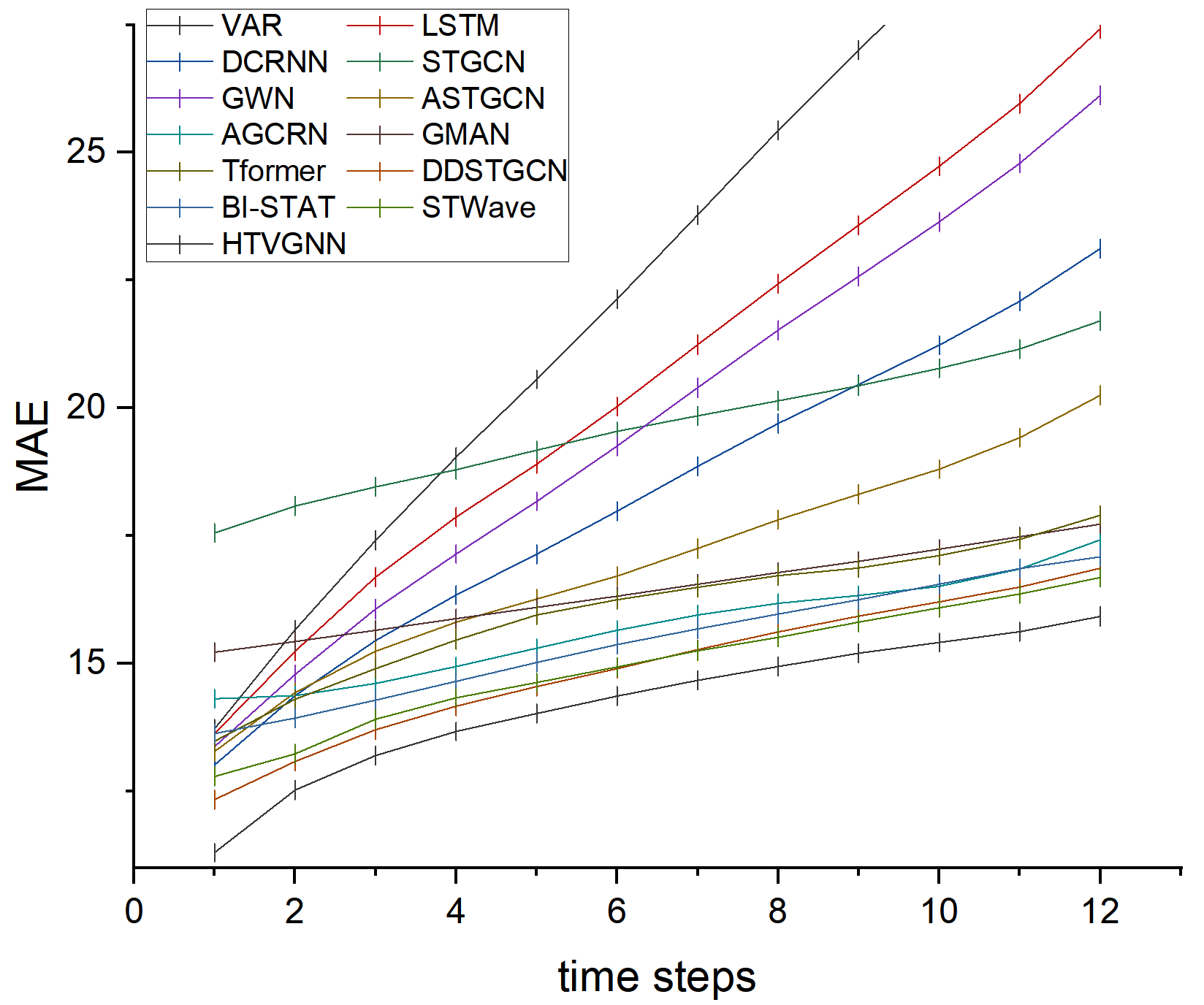}
         \caption{MAE on PEMS03}
         \label{fig:y equals x}
     \end{subfigure}
     \hfill
     \begin{subfigure}[b]{0.24\textwidth}
         \centering
         \includegraphics[width=\textwidth]{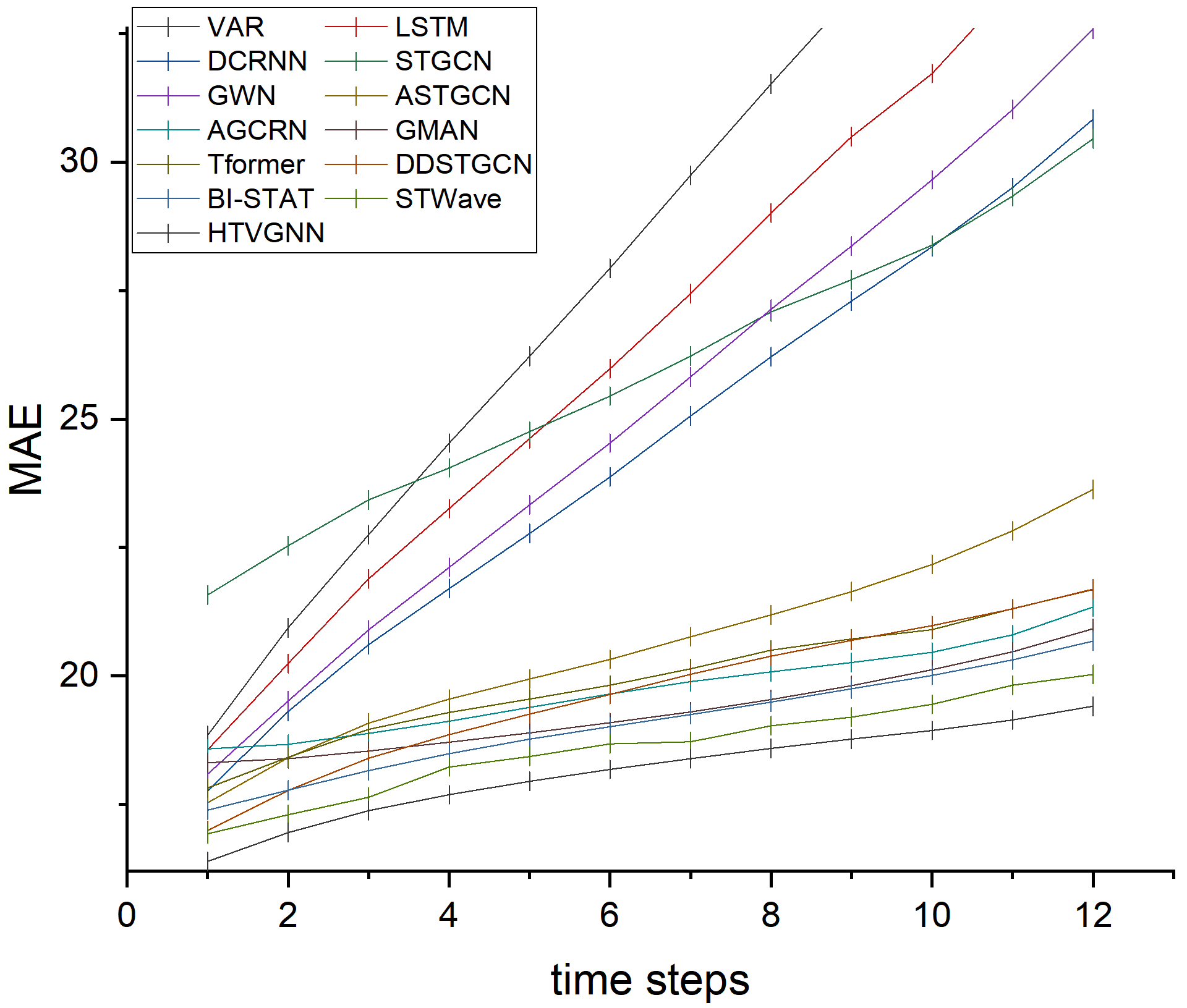}
         \caption{MAE on PEMS04}
         \label{fig:y equals x}
     \end{subfigure}
     \hfill
     \begin{subfigure}[b]{0.24\textwidth}
         \centering
         \includegraphics[width=\textwidth]{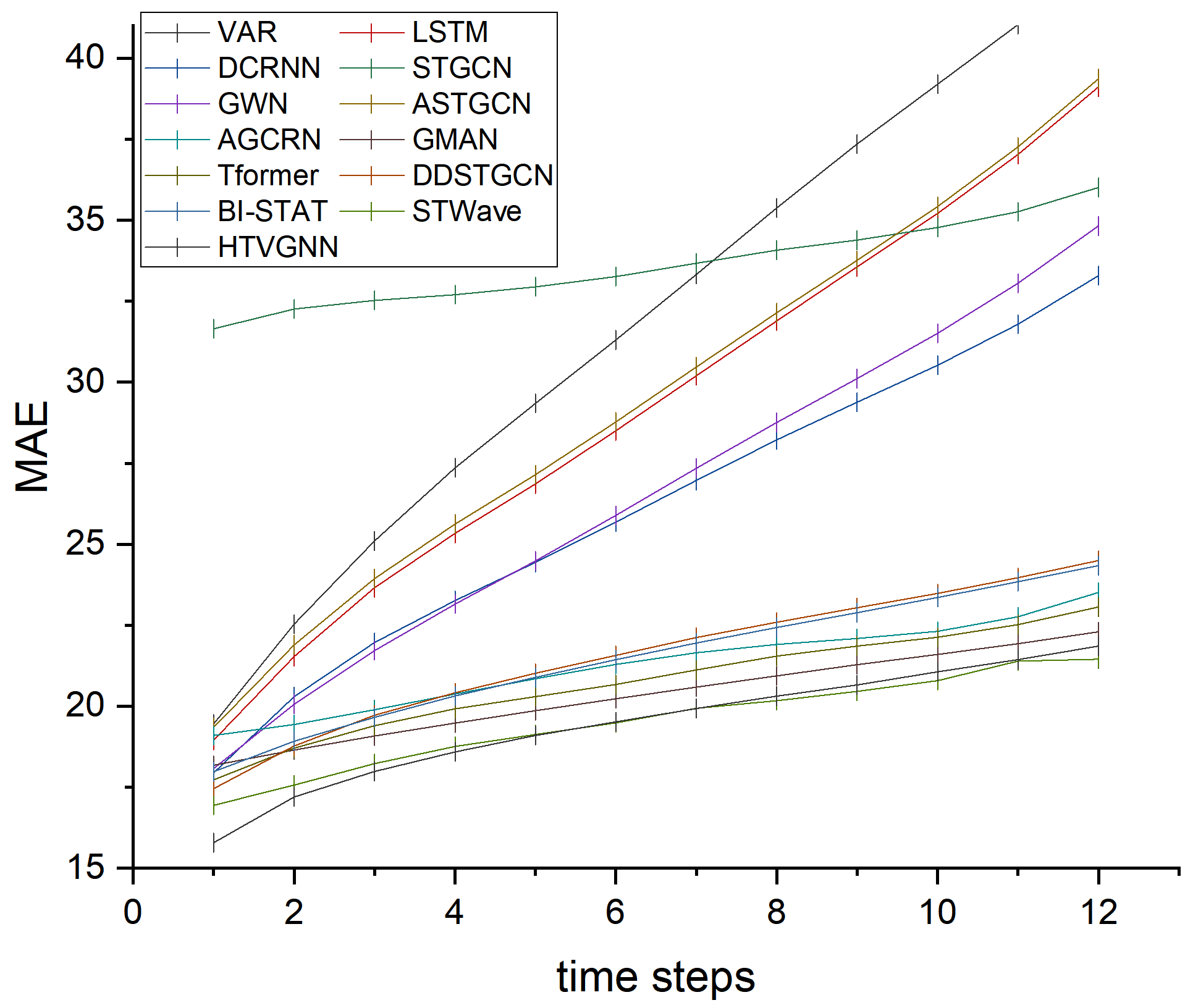}
         \caption{MAE on PEMS07}
         \label{fig:three sin x}
     \end{subfigure}
        \begin{subfigure}[b]{0.24\textwidth}
         \centering
         \includegraphics[width=\textwidth]{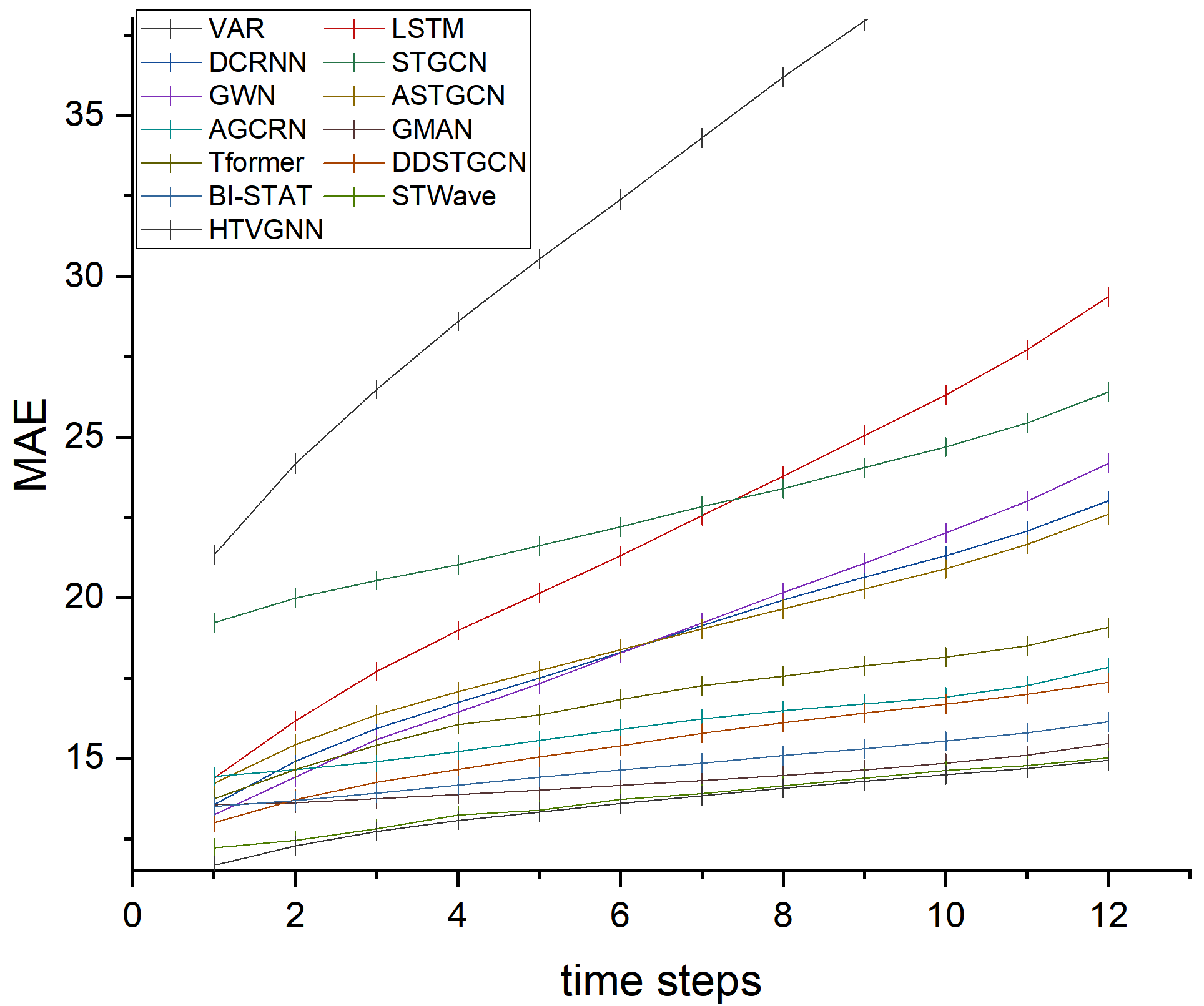}
         \caption{MAE on PEMS08}
         \label{fig:y equals x}
     \end{subfigure}
     \hfill
     \begin{subfigure}[b]{0.24\textwidth}
         \centering
         \includegraphics[width=\textwidth]{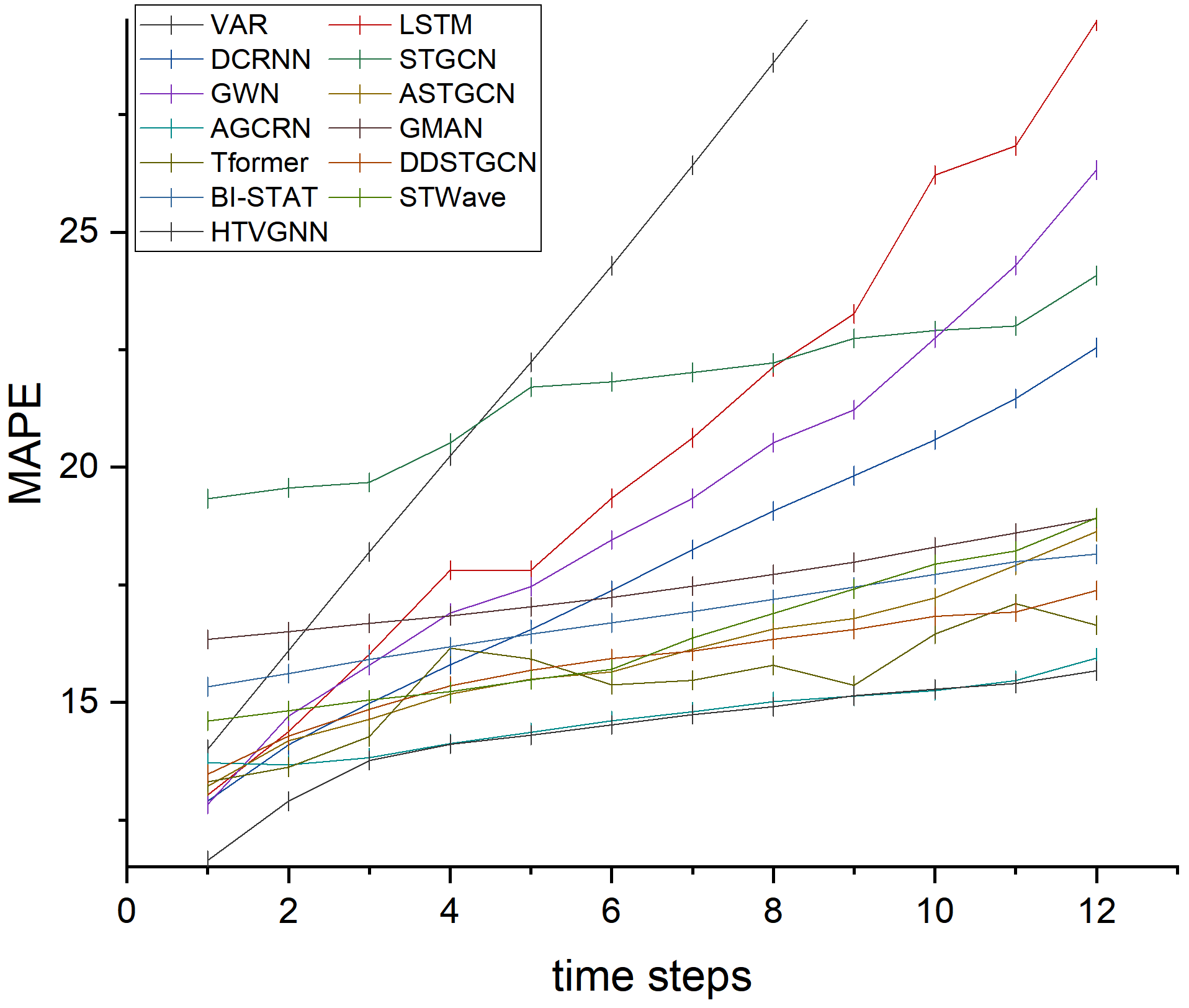}
         \caption{MAPE(\%) on PEMS03}
         \label{fig:y equals x}
     \end{subfigure}
     \hfill
     \begin{subfigure}[b]{0.24\textwidth}
         \centering
         \includegraphics[width=\textwidth]{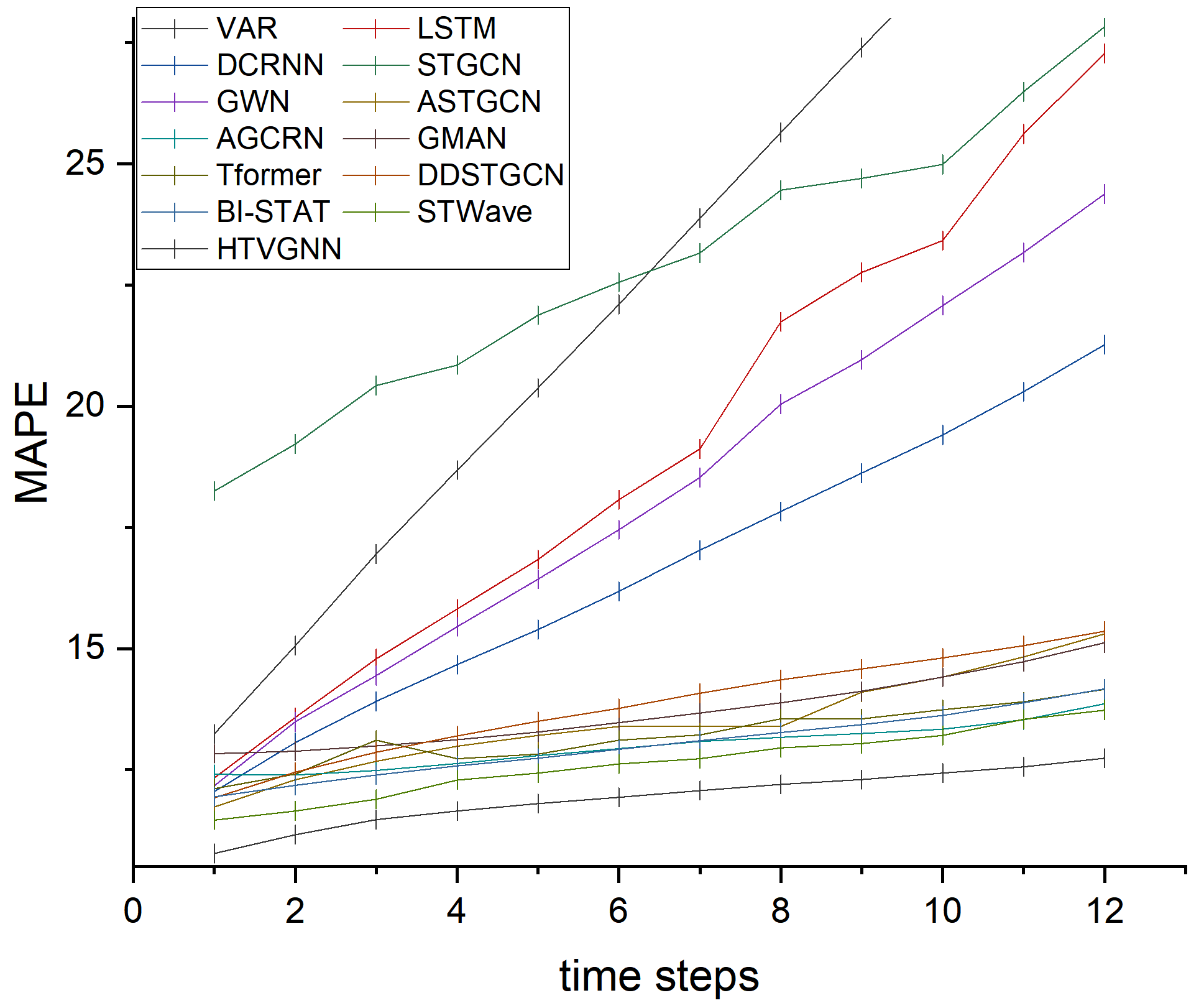}
         \caption{MAPE(\%)  on PEMS04}
         \label{fig:three sin x}
     \end{subfigure}
     \hfill
        \begin{subfigure}[b]{0.24\textwidth}
         \centering
         \includegraphics[width=\textwidth]{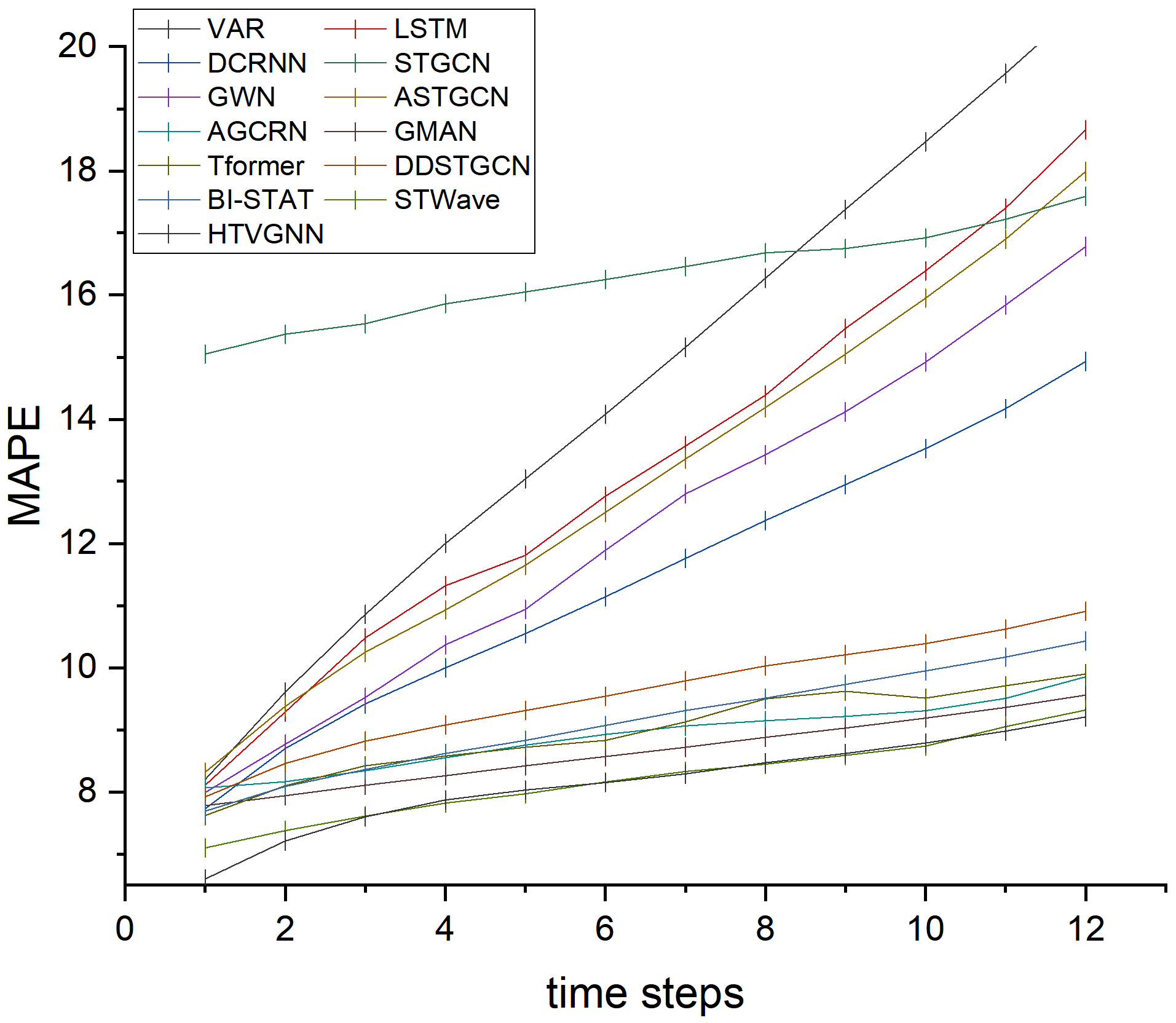}
         \caption{MAPE(\%) on PEMS07}
         \label{fig:y equals x}
     \end{subfigure}
     \hfill
     \begin{subfigure}[b]{0.24\textwidth}
         \centering
         \includegraphics[width=\textwidth]{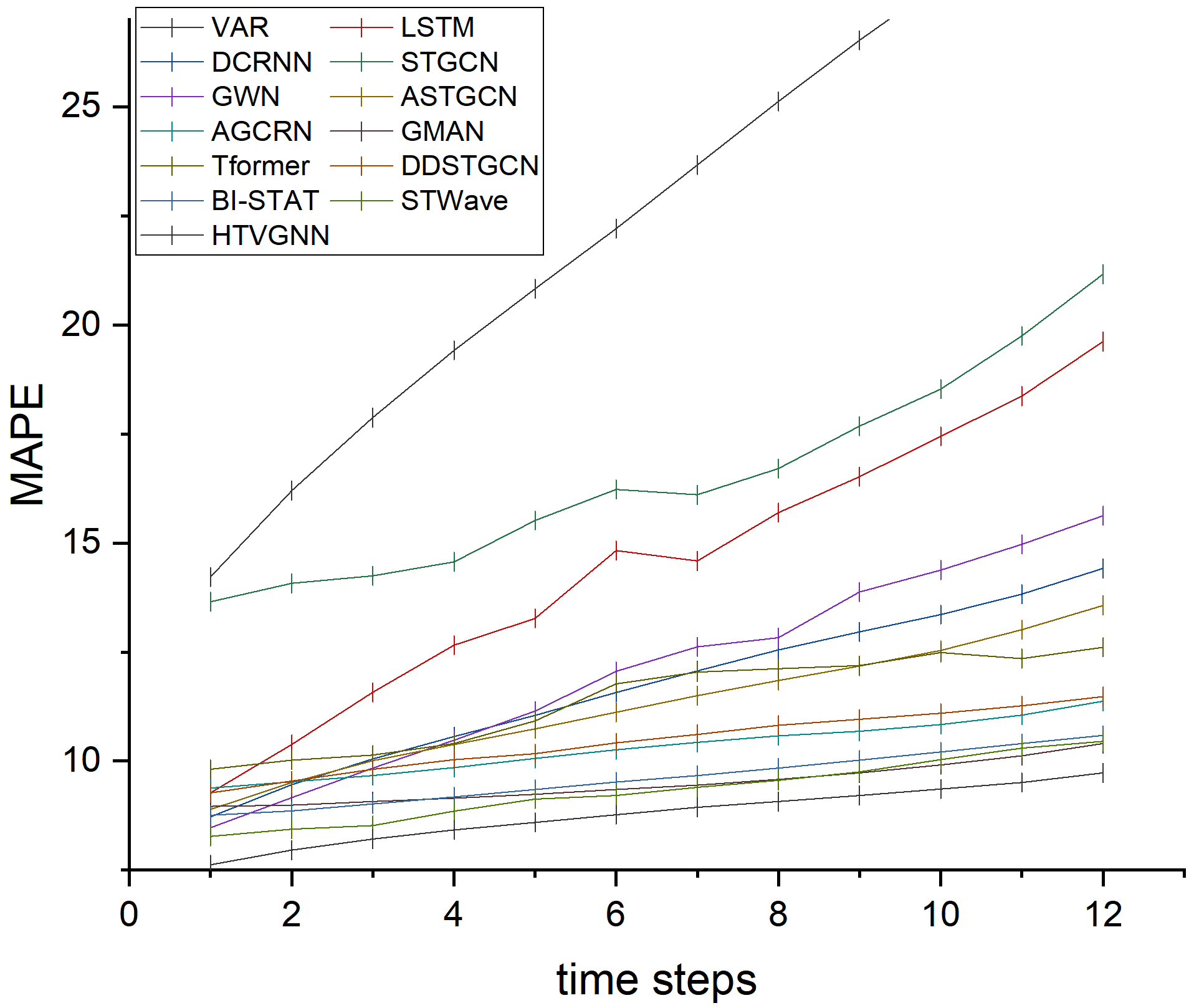}
         \caption{MAPE(\%) on PEMS08}
         \label{fig:y equals x}
     \end{subfigure}
     \hfill
     \begin{subfigure}[b]{0.24\textwidth}
         \centering
         \includegraphics[width=\textwidth]{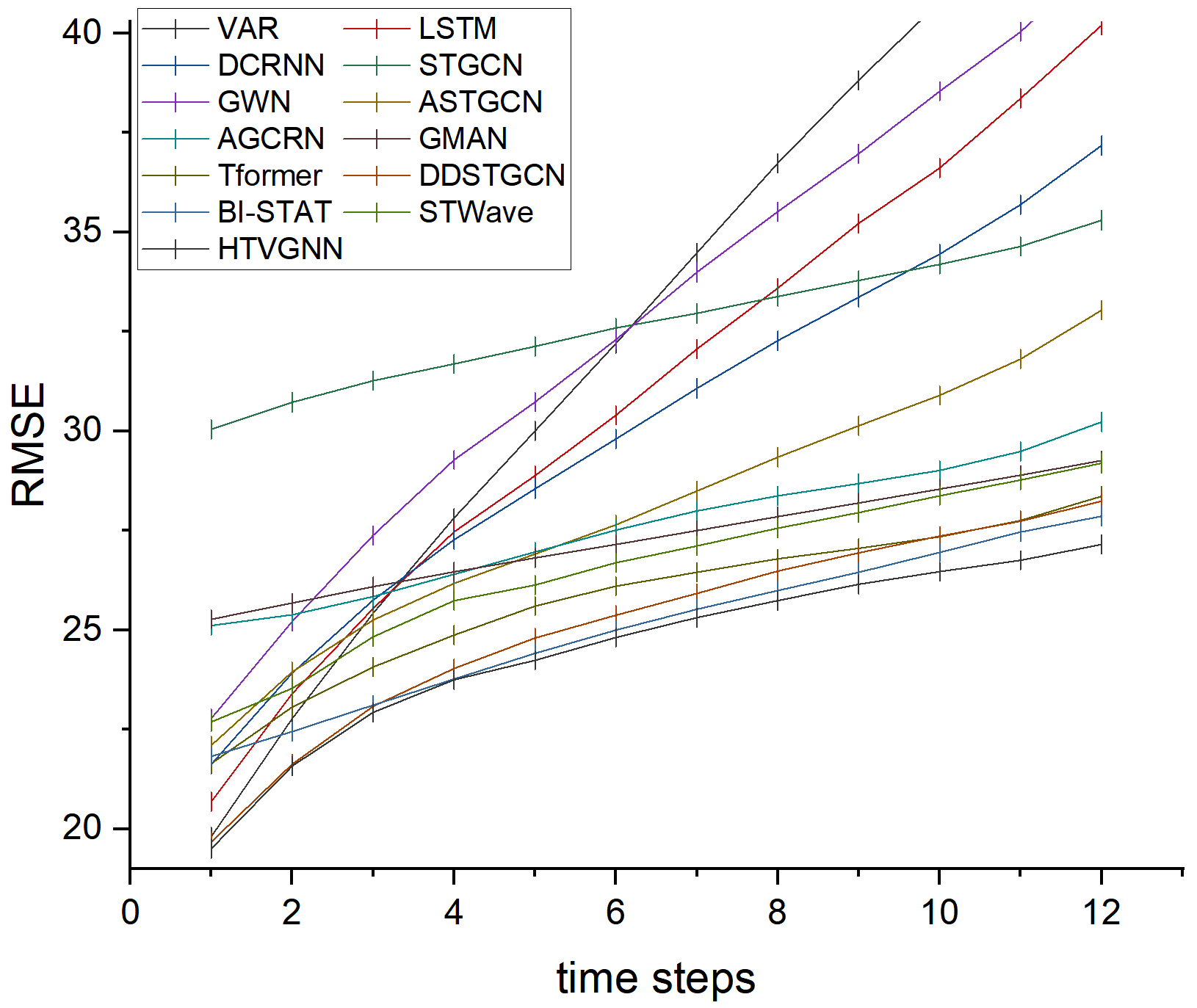}
         \caption{RMSE on PEMS03}
         \label{fig:three sin x}
     \end{subfigure}
     \hfill
     \begin{subfigure}[b]{0.24\textwidth}
         \centering
         \includegraphics[width=\textwidth]{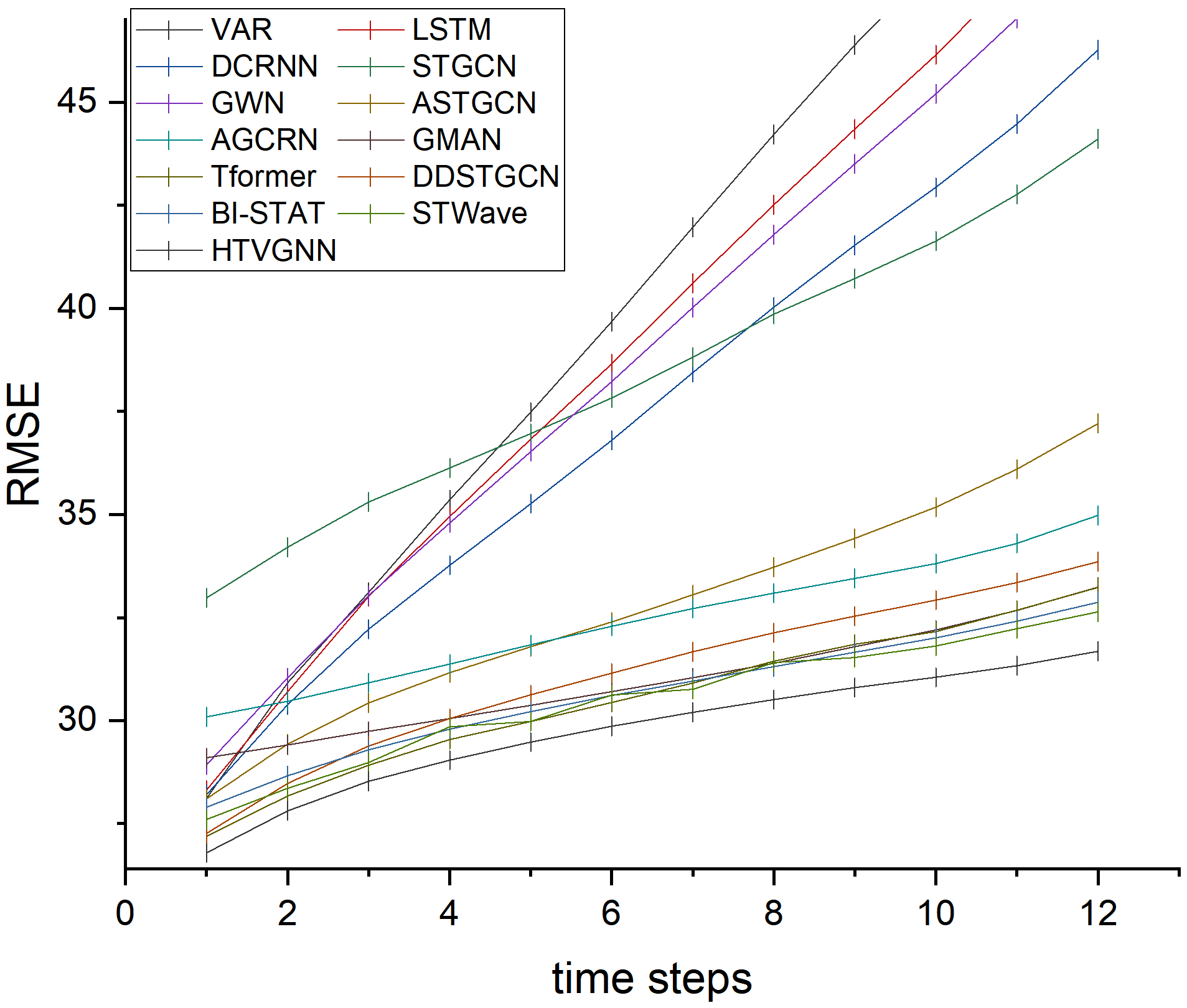}
         \caption{RMSE on PEMS04}
         \label{fig:y equals x}
     \end{subfigure}
     \hfill
     \begin{subfigure}[b]{0.24\textwidth}
         \centering
         \includegraphics[width=\textwidth]{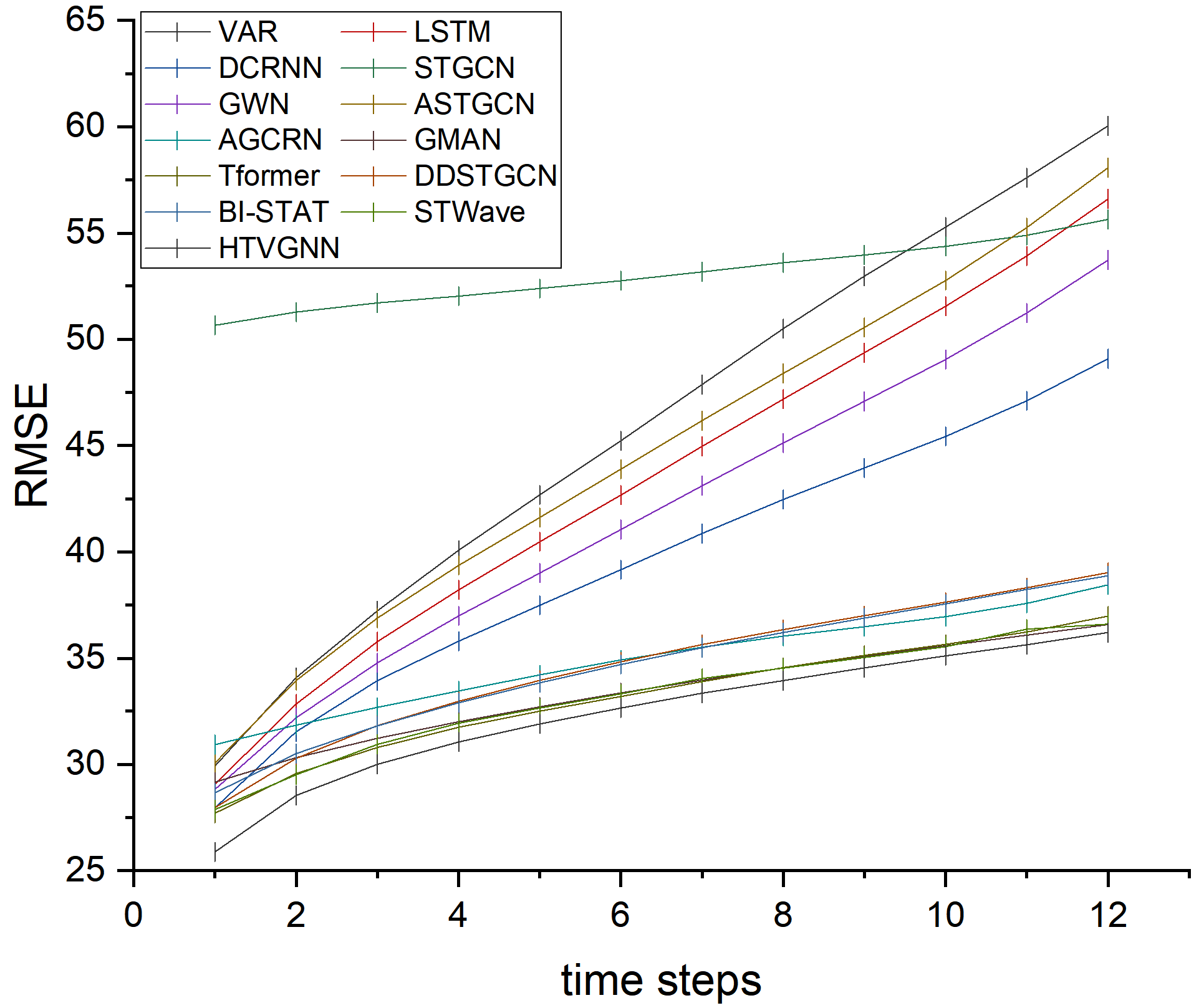}
         \caption{RMSE on PEMS07}
         \label{fig:y equals x}
     \end{subfigure}
     \hfill
     \begin{subfigure}[b]{0.24\textwidth}
         \centering
         \includegraphics[width=\textwidth]{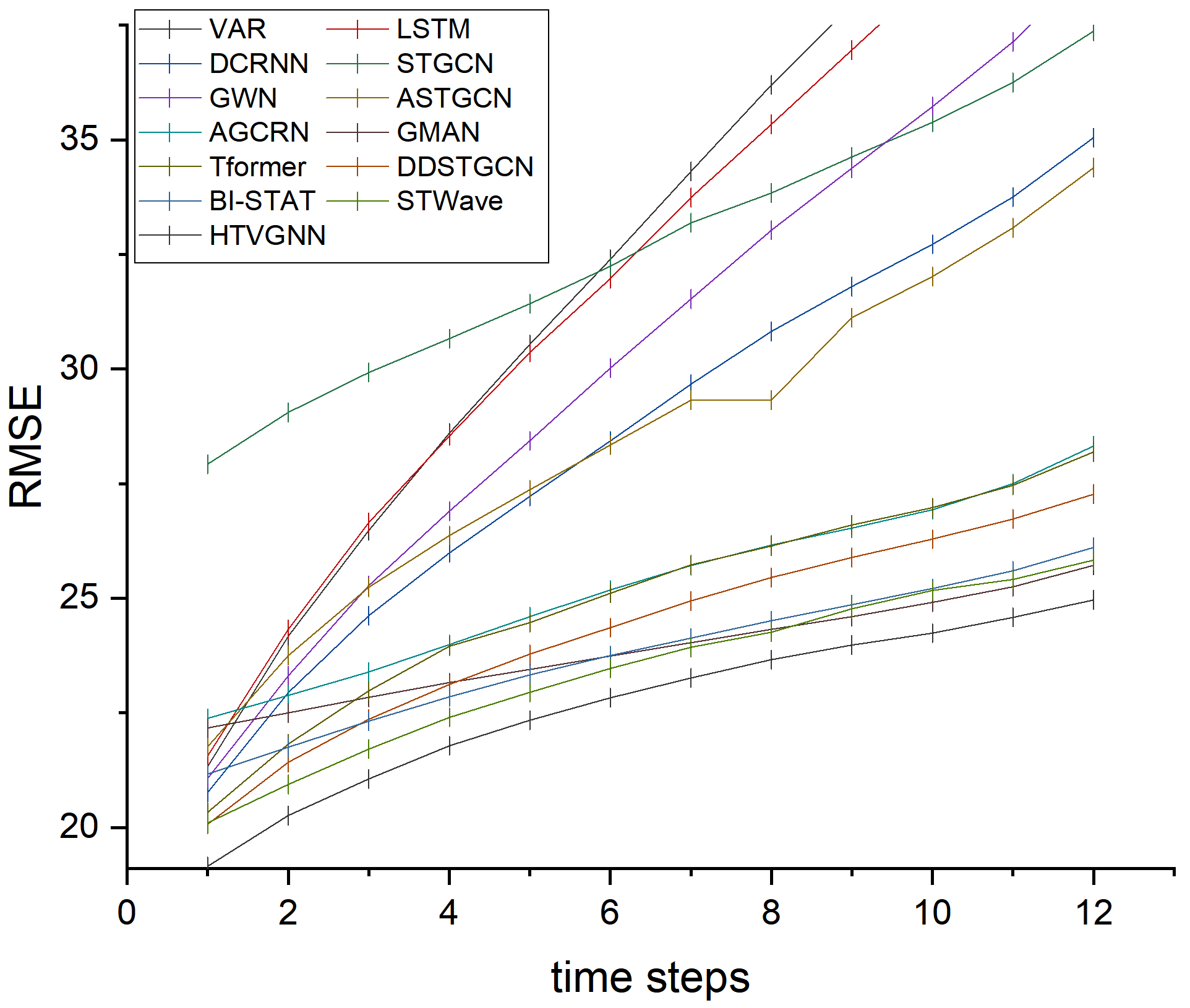}
         \caption{RMSE on PEMS08}
         \label{fig:three sin x}
     \end{subfigure}
     \hfill
     \caption{Prediction error of the different models on the four datasets at the different time step}
        \label{fg:5}
\end{figure*}
\begin{table*}[htbp]
\tabcolsep=0.15cm
\renewcommand\arraystretch{1.2}
\centering
\caption{Hyperparameters of HTVGNN model}
\begin{tabular}{ccccccccc}
\hline
Data & Encoder layers & Decoder layers &Batchsize& Heads & D & E & $d_m$ & CTVGCRM layers\\ \hline
PEMS03 & 1 & 1& 16& 8& 64 & 8 & 8 & 2\\
PEMS04 & 1 & 1& 4 & 8 & 64& 6 & 18 & 2\\
 PEMS07 & 1 & 1& 8& 8 & 64& 10 & 24 & 2\\
 PEMS08 & 1 & 1& 16& 8 & 64 & 5 & 15 & 2\\  \hline
\end{tabular}\label{tb:4-1}
\end{table*}
\subsection{Hyperparameter setting}
\subsubsection{Dataset Preprocessing}
Following recent methodologies \cite{ref31}, \cite{ref32} and  \cite{ref33}, we partition the training, validation, and test sets in a 6:2:2 ratio. To ensure more stable training, we employ z-score standardization. Additionally, we utilize historical traffic flow speed data from the past $T=12$ time slices (60 min) to predict future traffic flow speed in $\tau = 12$ time slices (60 min). The input data is normalized using Z-Score as shown in equation (\ref{eq0019}).
%
%
\begin{equation}\label{eq0019}
   \widehat x\  =  \ \frac{x\ \ -\ \ mean(x_{trian})}{std(x_{train})}
\end{equation}
%
%
Among them, mean() and std() represent the mean and standard deviation functions, and $x_{trian}$ represents the training set.
\subsubsection{Model Settings}
The experiments are conducted on a machine equipped with a single NVIDIA RTX3090 and 32GB of memory. HTVGNN is trained using Ubuntu 18.04, PyTorch 2.0.0, and the Adam optimizer \cite{ref34}. The learning rate is set to 0.001, while the remaining hyperparameters of the model are presented in the table~\ref{tb:4-1}.
\subsection{Performance Comparison}
Tables \ref{tb:2} and \ref{tb:3} present the comparison results of 12 prediction models on four datasets. Overall, our HTVGNN achieved exceptional prediction accuracy across all 12 prediction steps. The deep learning-based approaches outperformed traditional statistical-based approaches (VAR), indicating their efficacy in modeling highly nonlinear traffic flows. As depicted in Tables \ref{tb:2} and \ref{tb:3}, these approaches exhibited suboptimal prediction performance. Moreover, methods that capture spatio-temporal dependencies generally surpassed those solely modeling temporal dependencies (LSTM), underscoring the significance of spatial dependencies. DCRNN and STGCN employ predefined graphs to model spatial dependencies in road networks; however, their performance heavily relies on the quality of these predefined graphs. AGCRN utilizes adaptive graphs to capture spatial dependencies, which typically enhances predictions compared to predefined graphs. Nevertheless, both predefined and adaptive graphs remain static during the prediction stage, disregarding dynamic changes in the road network. In contrast, models that dynamically model spatial dependencies (GMAN, STWave) significantly outperform other approaches based on static graph structures (DCRNN, STGCN, GWN). This can be attributed to their ability to learn dynamically changing graph structures and provide a broader representation space for modeling more complex spatio-temporal dependencies.Compared to other models, HTVGNN incorporates a time-varying mask matrix to enhance the multi-head attention mechanism for accurately capturing dynamic temporal correlations. Moreover, HTVGNN employs a dedicated coupled graph learning method tailored for multi-head attention, effectively improving its performance in long-term prediction tasks. Additionally, to better model dynamic spatial correlations in road networks, we define a mask based on topological and semantic matrices to optimize the dynamic time-varying graphs. Overall, our model outperforms existing state-of-the-art approaches on four datasets, achieving superior prediction accuracy.

Additionally, in order to demonstrate the overall superiority of our model, we provide visual representations for each prediction point of HTVGNN and $13$ comparative models as depicted in Figure \ref{fg:5}. In terms of the PEMS03 and PEMS04 datasets, our model exhibits the smallest cumulative error as the increase of the length of time steps, highlighting a distinct advantage of HTVGNN during long-term prediction stages. Furthermore, for PEMS07 and PEMS08 datasets, our model consistently maintains a significant advantage across all time steps.

\subsection{Abliation study}
The purpose of this section is to demonstrate the efficacy of the key components in our model through a series of experiments. The subsequent variations aim to indicate the effectiveness of different combinations of modules:

 $\mathbf{w/o \ ETPMSA}$: CTVGCRM should be utilized exclusively, with the exclusion of the enhanced temporal perception multi-head self-attention mechanism. This variant primarily demonstrates the function of the enhanced temporal perception multi-head self-attention.

$\mathbf{w/o \ TV}$: Replace the time-varying graph in CTVGCRM with a topological graph, emphasizing its pivotal role in capturing spatial correlations.

 $\mathbf{w/o \ TR}$: Replace CTVGCRM with a topological graph-based graph convolution, which primarily emphasizes the role of the RNN module in HTVGNN. At this juncture, the number of layers remains unchanged.

The experiments were conducted multiple times, and the optimal MAE, RMSE, and MAPE values for each variant on PEMS04 and PEMS08 are presented in Table \ref{tb:4}.
%
%
\begin{table}[htbp]
\tabcolsep=0.1cm
\renewcommand\arraystretch{1.1}
\centering
\caption{Ablation experiments on PEMS04 and PEMS08.}
\begin{tabular}{ccccc}
\hline
PEMS04 & HTVGNN & w/o ETPMSA & w/o TV & w/o TR\\ \hline
MAE & 17.99 & 18.75& 19.35& 18.48\\
RMSE & 29.74 & 30.55 & 31.29& 30.08 \\
 MAPE & 11.90 & 12.69 & 12.88& 12.39\\ \hline                   PEMS08 & HTVGCN & w/o ETPMSA & w/o TV & w/o TR\\ \hline
MAE & 13.24 & 14.58 & 15.74& 13.99 \\
RMSE & 22.67 & 23.66 & 25.06& 23.23 \\
 MAPE & 8.63 & 9.40 & 12.88& 9.07 \\ \hline
\end{tabular}\label{tb:4}
\end{table}
\begin{table*}[htbp]
\centering
\renewcommand\arraystretch{1.1}
\tabcolsep=0.08cm
\caption{Prediction performance of Transformer-o and Transformer-tv on datasets PEMS04 and PEMS08}\label{tb:5}
\begin{tabular}{cccccccccc}     \hline
\multirow{2}*{\# Layer}   & \multicolumn{3}{c}{\multirow{2}*{Variants}}  & \multicolumn{3}{c}{PEMS04} & \multicolumn{3}{c}{PEMS08}  \\ \cline{5-10}
                        & \multicolumn{3}{c}{}                       & MAE       & RMSE &MAPE(\%)       & MAE      & RMSE      &MAPE(\%)              \\ \hline
 \multirow{2}*{L=1} & \multicolumn{3}{c}{Transformer-o}                     & 19.52 & 31.58 & 13.38& 15.5 & 25.16 & 12.05 \\
                        & \multicolumn{3}{c}{Transformer-tv}                   & 18.89 & 30.52 & 12.76 & 14.80 & 23.98 & 9.54 \\ \hline

\multirow{2}*{L=4} & \multicolumn{3}{c}{Transformer-o}                     & 18.56 & 30.90 & 12.35& 14.83 & 24.43 & 9.45 \\
                       & \multicolumn{3}{c}{Transformer-tv} &18.45 & 30.16 & 12.13 & 13.75 & 23.28 & 8.98 \\ \hline

\end{tabular}
\end{table*}
%
%
%
%

The results presented in Table \ref{tb:4} demonstrate the superior performance of HTVGNN with ETPMSA, highlighting the crucial role played by ETPMSA in HTVGNN. Furthermore, the inferior performance of w/o TV compared to HTVGNN indicates that capturing spatial correlations through time-varying graphs is essential. Additionally, w/o TR being inferior to HTVGNN reveals the positive contribution of RNN in capturing temporal correlations.
%
%

Given that our model is a hybrid time-varying graph convolutional neural network, this tutorial aims to dissect the impact of time-varying mask embedding and time-varying embedding on its performance by dividing the model into modules, ultimately analyzing their influence on the model's performance.
%
%
\begin{figure*}
     \centering
     \begin{subfigure}[b]{0.45\textwidth}
         \centering
         \includegraphics[width=\textwidth]{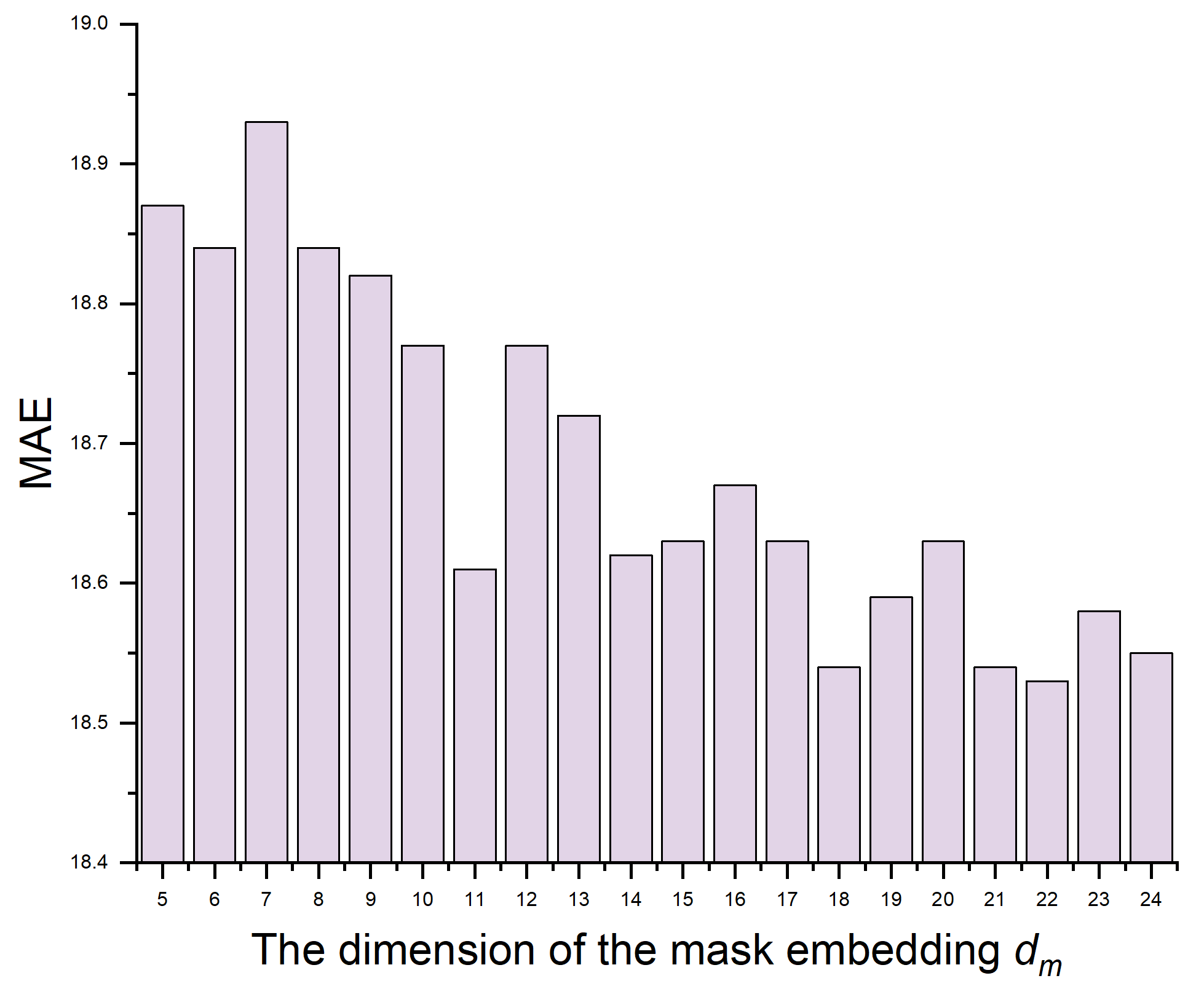}
         \caption{MAE on PEMS04}
         \label{fig:y equals x}
     \end{subfigure}
     \hfill
     \begin{subfigure}[b]{0.45\textwidth}
         \centering
             \includegraphics[width=\textwidth]{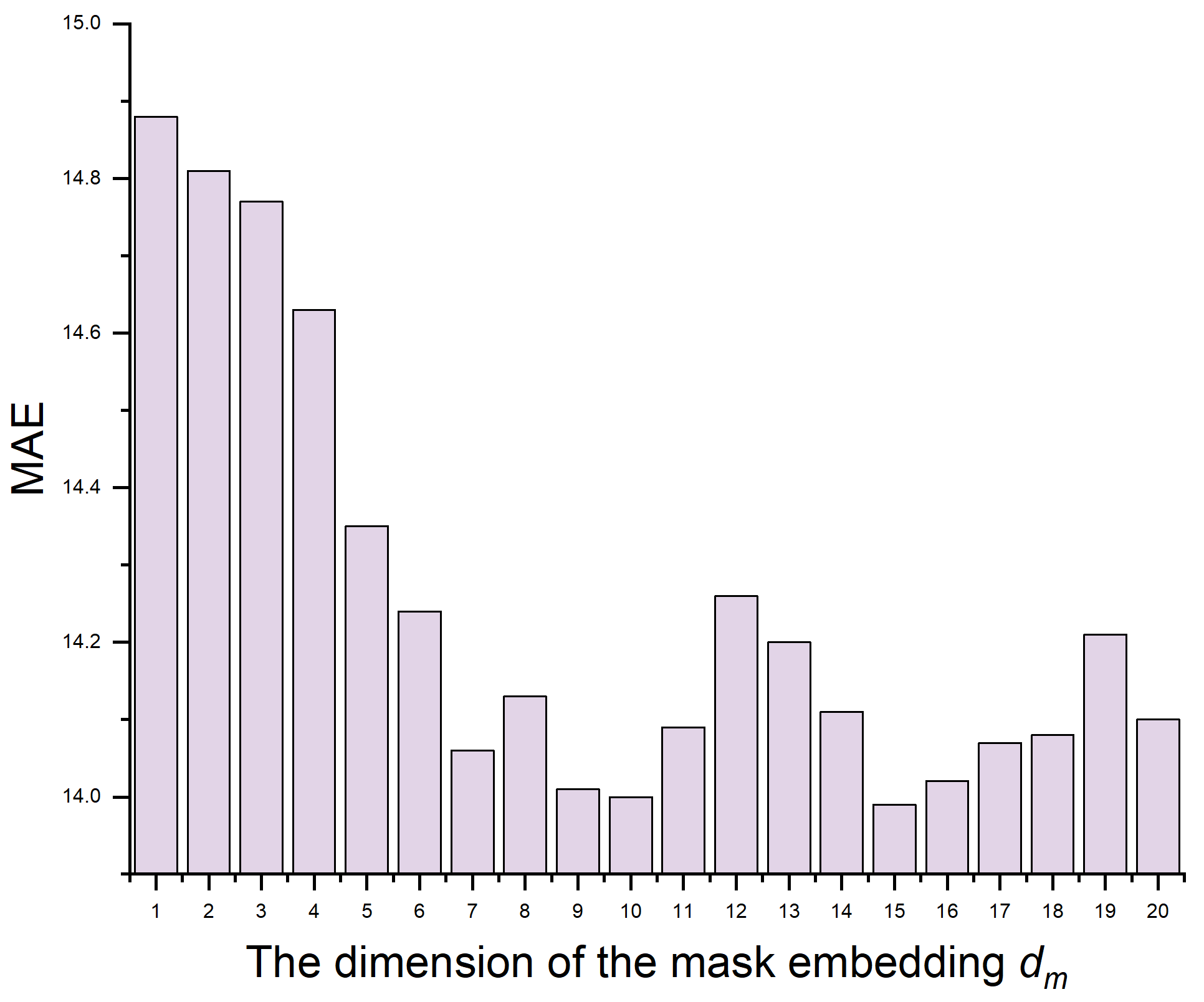}
         \caption{MAE on PEMS08}
         \label{fig:y equals x}
     \end{subfigure}
     \hfill
     \begin{subfigure}[b]{0.45\textwidth}
         \centering
         \includegraphics[width=\textwidth]{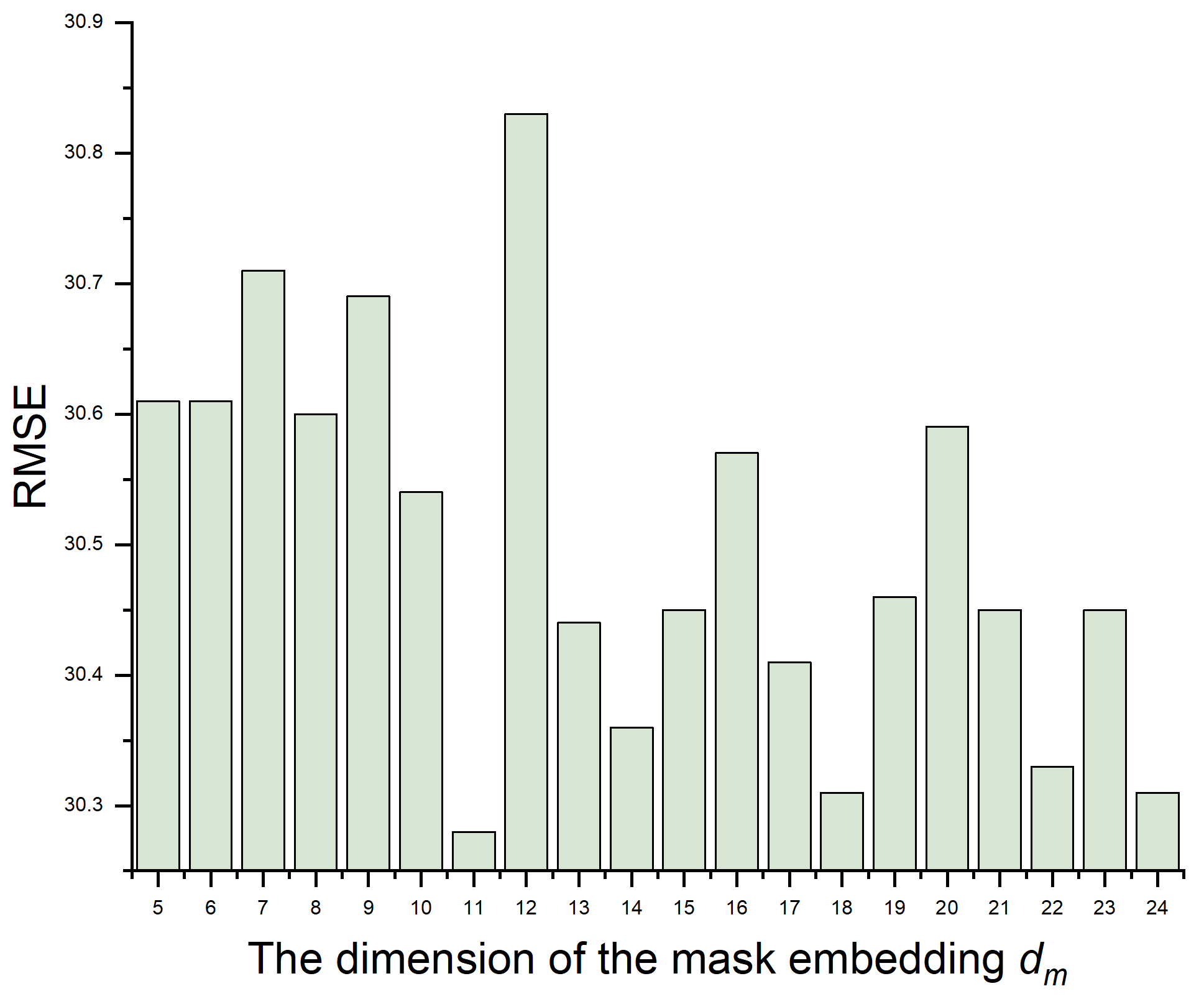}
         \caption{RMSE on PEMS04}
         \label{fig:three sin x}
     \end{subfigure}
     \hfill
     \begin{subfigure}[b]{0.45\textwidth}
         \centering
         \includegraphics[width=\textwidth]{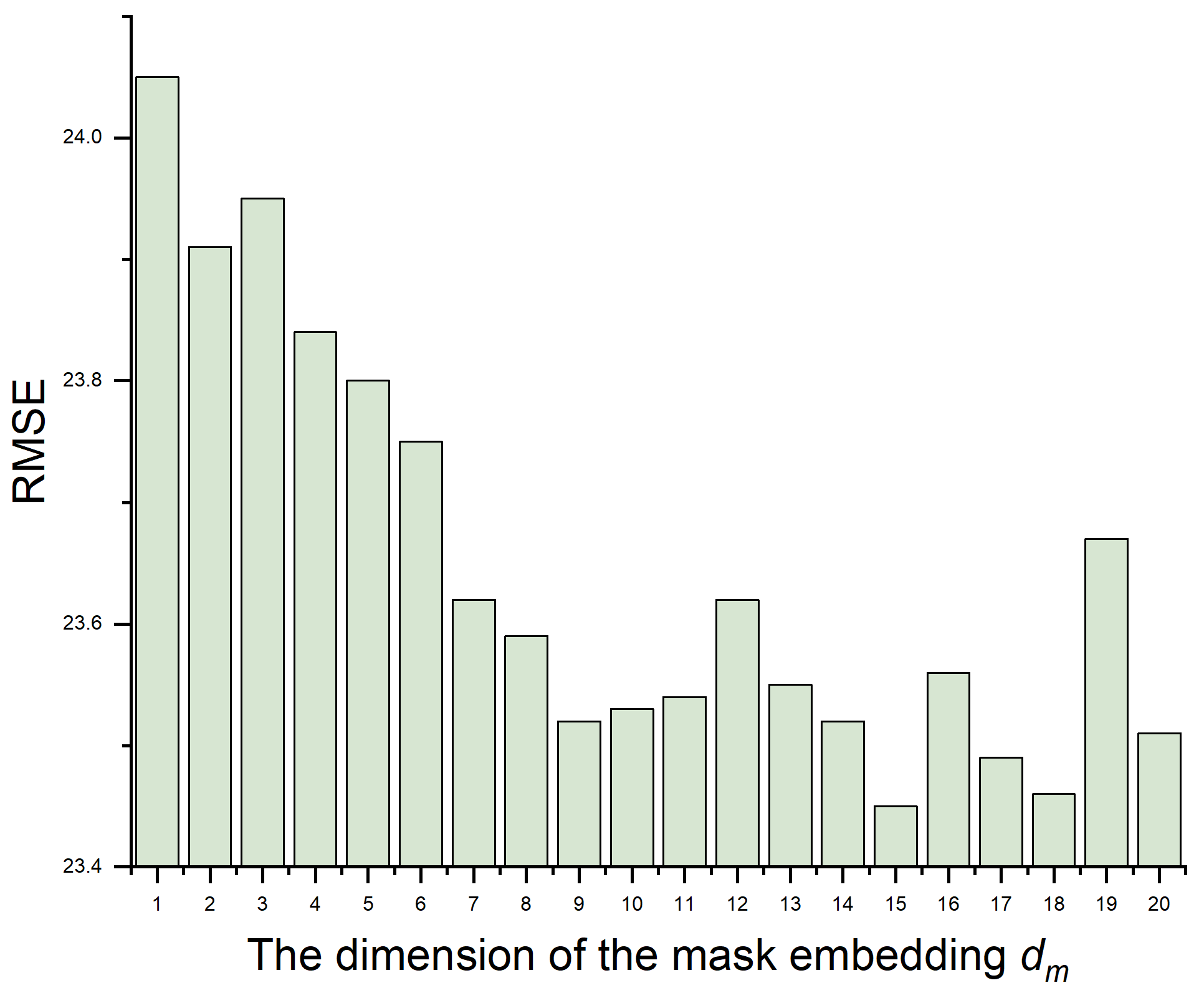}
         \caption{RMSE on PEMS08}
         \label{fig:y equals x}
     \end{subfigure}
     \hfill
     \caption{Prediction error of HTVGNN model on PEMS04 and PEMS08 datasets with different mask embeddings}
        \label{fg:6}
\end{figure*}
%
%
\subsubsection{Effect analysis of time-varying mask embedding in enhanced temporal perception multi-head attention}  
%
%
In order to demonstrate the effectiveness of our proposed time-varying mask embedding in enhanced temporal perception multi-head attention, the Transformer model with temporal trend-aware multi-head self-attention reported in \cite{GuoSN2021} is selected as the comparison model and denoted as $Transformer-o$. For ease of description,  the enhanced temporal perception multi-head attention with time-varying mask embedding is denoted as $Transformer-tv$. 
Firstly, we investigate the impact of dimension $d_{m}$ of the mask embedding $\tilde E \in R^{N \times d_m}$, $\tilde P_t^D \in R^{N \times d_m}$ and  $\tilde P^W_t \in R^{N \times d_m}$ in the enhanced temporal perception multi-head self-attention. For the ease of analysis, we set the number of layers of $Transformer-o$ and that of $Transformer-tv$ equal to $1$. Then, we can obtain the optimal mask embedding. 
Figure \ref{fg:6} illustrates how different time-varing mask embeddings affect PEMS04 and PEMS08 datasets. As depicted in Figure \ref{fg:6}, the optimal mask embeddings for PEMS04 and PEMS08 are respectively $18$ and $15$.


To highlight the advantages of enhanced temporal perception in multi-head self-attention, we set the number of layers to $4$ based on optimal mask embedding.
%
%
%
%
\begin{table}[htbp]
\tabcolsep=0.2cm
\renewcommand\arraystretch{1.1}
\centering
\caption{The impact of different graphs structures.}
\begin{tabular}{ccccc}
\hline
Dataset & Graphs structures & MAE & RMSE & MAPE(\%)\\ \hline
\multirow{4}*{PEMS04}
 & CTVGCRM & 18.76 & 30.55 & 12.69 \\
 & CTVGCRM-sl & 18.93 & 31.05 & 12.73 \\
 & CTVGCRM-ag & 19.35 & 31.29 & 12.88 \\
 & CTVGCRM-s & 19.56 & 31.53 & 13.15  \\ \hline
 \multirow{4}*{PEMS08}
  & CTVGCRM & 14.58 & 23.66 & 9.40 \\
 & CTVGCRM-sl & 15.08 & 24.25 & 9.77 \\
 & CTVGCRM-ag & 15.24 & 24.38 & 10.66 \\
 & CTVGCRM-s & 15.74 & 25.06 & 9.87  \\\hline
\end{tabular}\label{tb:6}
\end{table}
%
%
\begin{table}[htbp]
\tabcolsep=0.15cm
\renewcommand\arraystretch{1.2}
\centering
\caption{Validation of the proposed enhanced temporal perception multi-head self-attention and coupling time-varying graphs.}
\begin{tabular}{ccccc}
\hline
PEMS04 & HTVGNN & w/o-tm & w/o-cg & w/o-bc \\ \hline
MAE & 17.99 & 18.53 & 18.35& 18.53\\
RMSE & 29.74 & 30.33 & 30.30& 30.52 \\
MAPE(\%) & 11.90 & 12.30 & 12.15& 12.45 \\ \hline
PEMS08 & HTVGNN & w/o-tm & w/o-cg & w/o-bc \\ \hline
MAE & 13.24 & 14.10 & 13.79& 14.55 \\
RMSE & 22.67 & 23.42 & 23.11& 23.75 \\
MAPE(\%) & 8.63 & 9.10 & 9.02& 9.30\\ \hline
\end{tabular}\label{tb:7}
\end{table}
%
%

\begin{table*}[htbp]
\centering
\renewcommand\arraystretch{1.2}
\tabcolsep=0.14cm
\caption{Validation of the effectiveness of coupled time-varying graphs in long-term prediction}
\begin{tabular}{ccccccccccccc}     \hline
\multirow{2}*{Dataset}   & \multicolumn{3}{c}{\multirow{2}*{Variants}}  & \multicolumn{3}{c}{15min} & \multicolumn{3}{c}{30min} & \multicolumn{3}{c}{60min}  \\ \cline{5-13}
                        & \multicolumn{3}{c}{}                       & MAE       & RMSE &MAPE(\%)       & MAE      & RMSE      &MAPE(\%)      & MAE      & RMSE      &MAPE(\%)           \\ \hline
\multirow{2}*{PEMS04} & \multicolumn{3}{c}{w/o-cg}                      & 17.49 & 28.67 & 11.55 & 18.59 & 30.49 & 12.19 & 20.17 & 32.96 & 13.29 \\
                        & \multicolumn{3}{c}{HTVGNN}                    &17.25      & 28.48 & 11.44        &   18.01       & 29.81   &  11.89    & 19.20       & 31.62 &  12.72\\ \hline

\multirow{2}*{PEMS08} & \multicolumn{3}{c}{w/o-cg}                      & 12.90 & 21.34 & 8.45 & 13.96     & 23.35  & 9.13      &    15.53     &  25.83 & 10.06 \\
                        & \multicolumn{3}{c}{HTVGNN}  &  12.41      & 21.06 & 8.09 & 13.28     & 22.83 & 8.65      &    14.48      & 24.96 & 9.44 \\ \hline

\end{tabular}\label{tb:8}
\end{table*}
%
%
%

The results of the experimental with different numbers of layers are presented in Table \ref{tb:5}. It is evident that $Transformer-tv$ outperforms $Transformer-o$ significantly when the number of layers is equal to $1$ and $4$. Our proposed time-varying mask matrix effectively mitigates the issue of inaccurate calculation in multi-head self-attention, while greatly enhancing temporal perception capability.
%
%
\subsubsection{Effect analysis of different types of graph structure in CTVGCRM and HTVGNN}

Similar to the preceding section, in order to evaluate the impact of diverse graph structures on CTVGCRM, we formulated four iterations of CTVGCRM as delineated follows:

$\textbf{CTVGCRM}$: This is a coupling time-varying graph convolution gated recurrent network.

$\textbf{CTVGCRM-sl}$: Let different time slices be learned with different embeddings, but the spatial correlations learned in different time slices are not coupled.

$\textbf{CTVGCRM-ag}$: All time slices are learned using the same embedding, in which case the time-varying graph is the same as AGCRN.

$\textbf{CTVGCRM-s}$: Replace coupled time-varying graphs with topological graphs.
%
%

The results presented in Table \ref{tb:6} unequivocally demonstrate the superiority of CTVGCRM over  its variants, namely CTVGCRM-sl ,CTVGCRM-ag and CTVGCRM-s.This highlights the limitations of predefined graphs in accurately representing spatial correlations on traffic networks. Furthermore, the superior performance of CTVGCRM-sl compared to CTVGCRM-ag suggests that spatial correlations vary across different time steps, indicating that employing a single time-varying graph for learning is suboptimal. Moreover, the superiority of CTVGCRM over CTVGCRM-sl implies that considering the corresponding spatial correlations for different time steps is crucial as it allows capturing relevant information from other time steps to enhance the modeling accuracy of the current time step.

%
%

%
%
%
%
%
%

%
%
Subsequently, we assess the impact of our proposed enhanced temporal perception multi-head self-attention and coupling time graph on the overall model performance by leveraging the aforementioned optimal mask embedding and dropout techniques. We devise a series of five variants to conduct ablation experiments on our model using PEMS04 and PEMS08 datasets. These variants aim to highlight the significance of time-varying mask embedding and time-varying embedding in our approach, as described below:
%
%

$\mathbf{HTVGNN}$: A novel hybrid time-varying graph neural network, as shown in Fig. \ref{fg:0001}

 $\mathbf{w/o-tm}$: Remove the time-varying mask embedding part of the enhanced temporal perception multi-head self-attention in HTVGNN.

$\mathbf{w/o-cg}$: Remove the coupling part of the time-varying graph in HTVGNN.

 $\mathbf{w/o-bc}$: Remove both the time-varying mask embedding part of the enhanced temporal perception multi-head self-attention and the coupling part of the time-varying graph from HTVGNN.

%
%

The results presented in Table \ref{tb:7} demonstrate that the superiority of HTVGNN over both w/o-tm, highlighting the positive impact of incorporating time-varying mask embedding on enhancing self-attention accuracy. Furthermore, the superior performance of HTVGNN compared to w/o-cg and the superior performance of w/o-cg compared to w/o-bc indicates its advantage in capturing spatial correlations in transportation networks by leveraging coupling time-varying graphs.

%
%
In particular, The ablation study revealed that our proposed HTVGNN, incorporating time-varying graphs, yields significant benefits in reducing the cumulative prediction error. Detailed comparison results are presented in Table \ref{tb:8}.

The results presented in Table \ref{tb:8} unequivocally demonstrate the superiority of HTVGNN with coupling time-varying graphs over its counterpart without such coupling. This advantage becomes increasingly pronounced as the time step increases. Furthermore, Table \ref{tb:8} reveals that HTVGNN outperforms HTVGNN-wc by 4.81\%, 4.06\%, and 4.29\% in terms of MAE, RMSE, and MAPE on PEMS04 dataset at a temporal resolution of 60 minutes, respectively. Similarly, on PEMS08 dataset at the same temporal resolution, HTVGNN achieves improvements of 6.76\%, 3.37\%, and 6.16\% for MAE, RMSE, and MAPE.

%
%
\begin{figure*}
     \centering
     \begin{subfigure}[b]{0.24\textwidth}
         \centering
         \includegraphics[width=\textwidth]{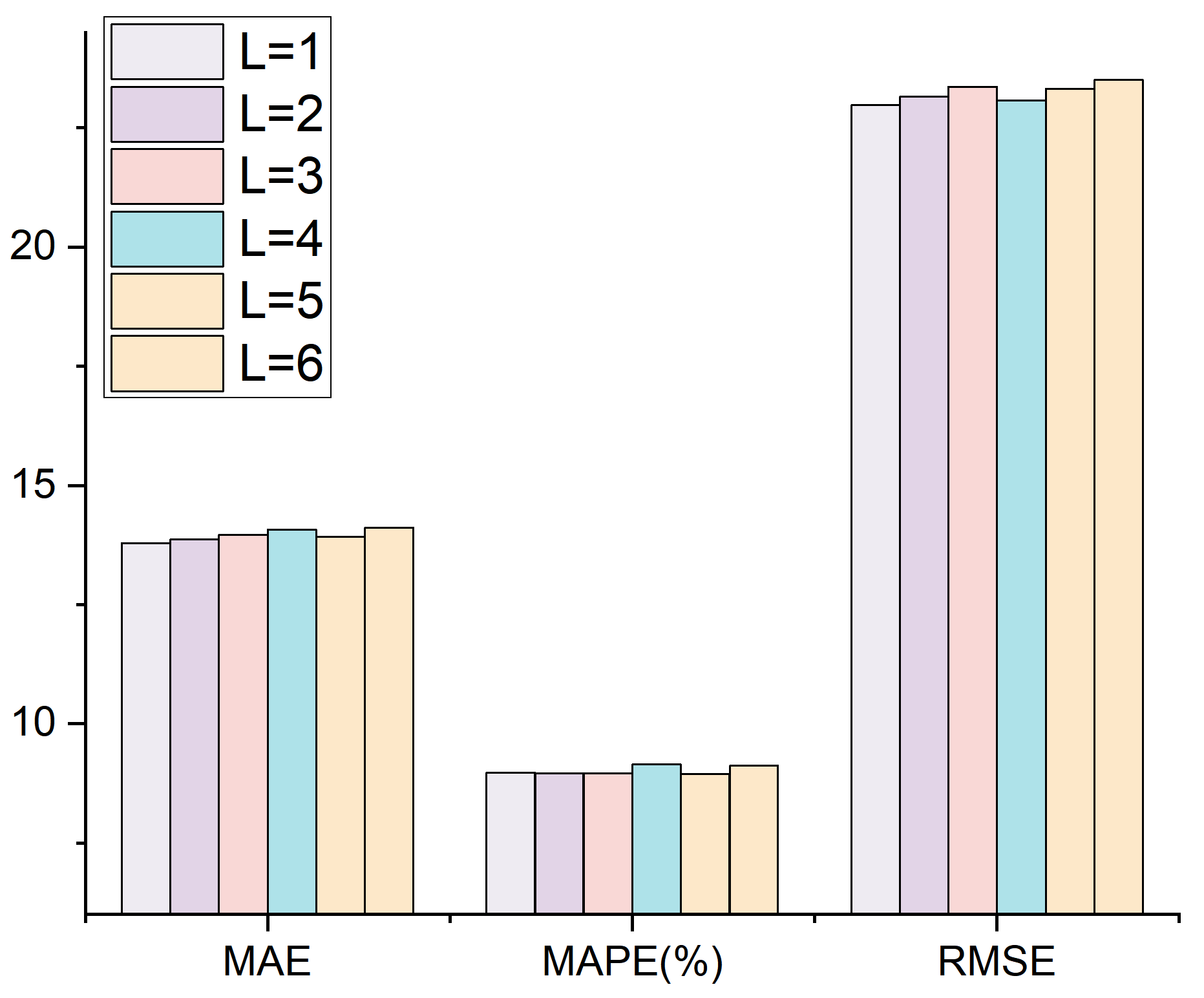}
         \caption{Sensitivity to L on PEMS08}
         \label{fig:5a}
     \end{subfigure}
     \hfill
     \begin{subfigure}[b]{0.24\textwidth}
         \centering
         \includegraphics[width=\textwidth]{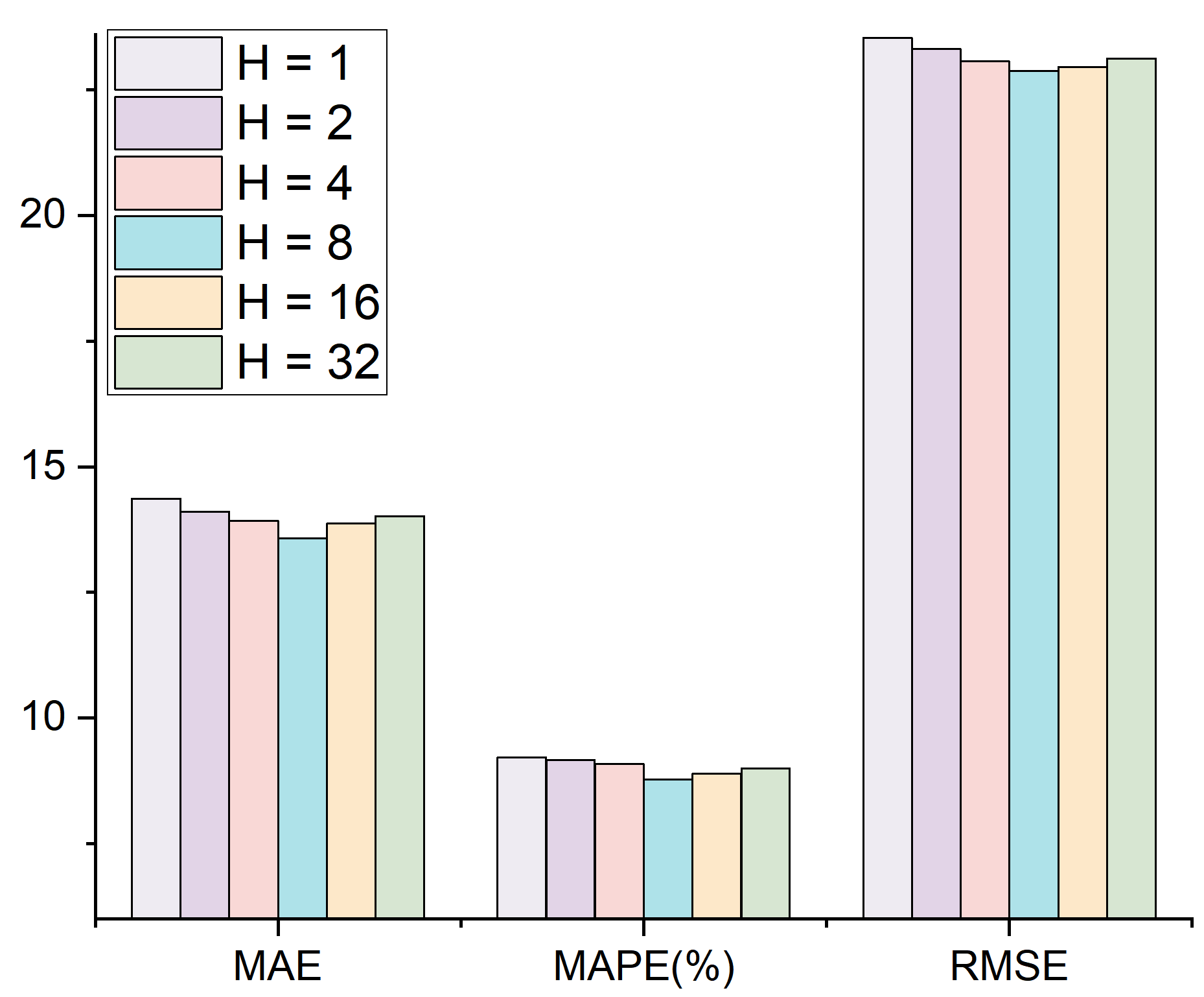}
         \caption{Sensitivity to H on PEMS08}
         \label{fig:5b}
     \end{subfigure}
     \hfill
        \begin{subfigure}[b]{0.24\textwidth}
         \centering
         \includegraphics[width=\textwidth]{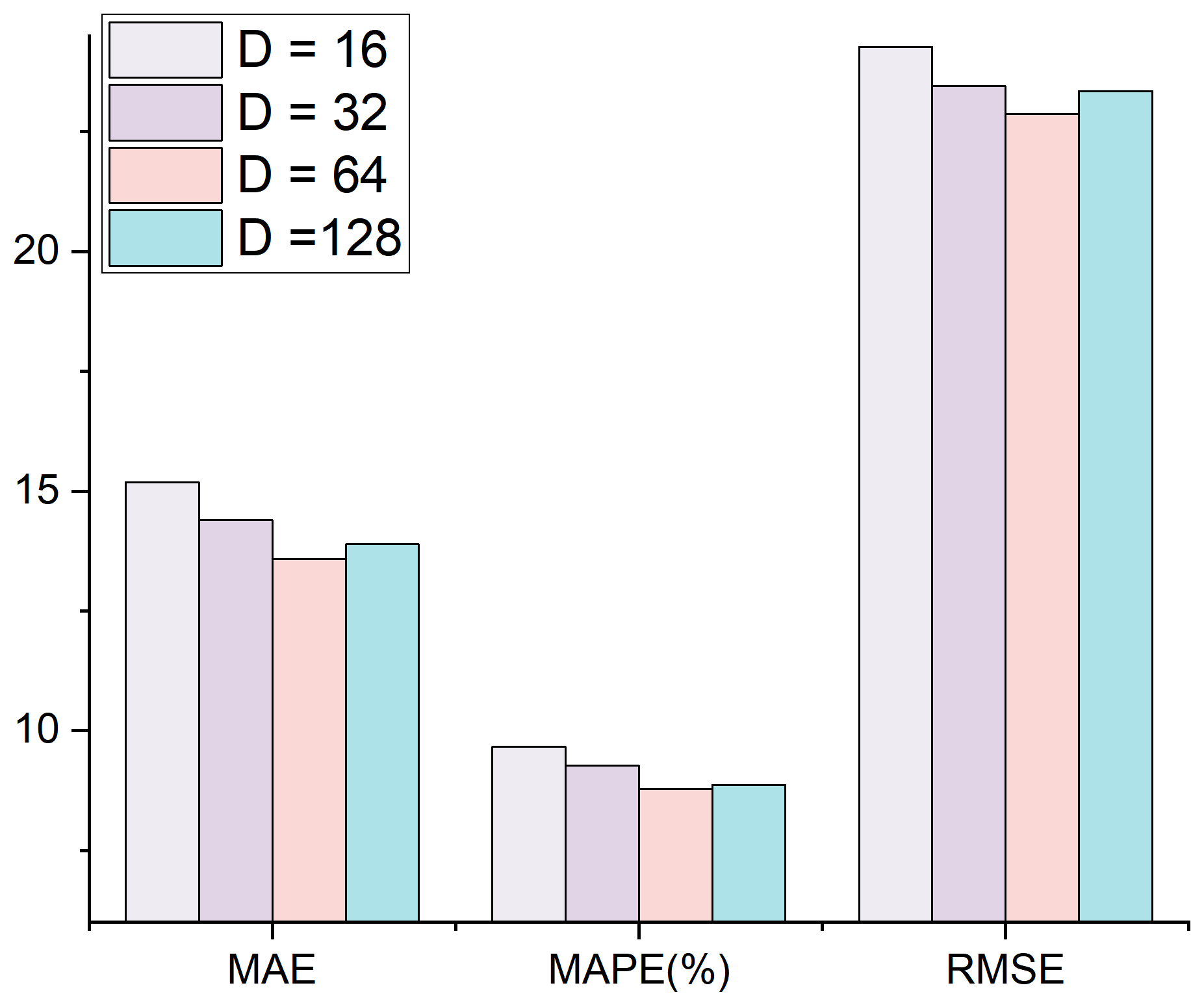}
         \caption{Sensitivity to D on PEMS08}
         \label{fig:5c}
     \end{subfigure}
     \hfill
     \begin{subfigure}[b]{0.24\textwidth}
         \centering
         \includegraphics[width=\textwidth]{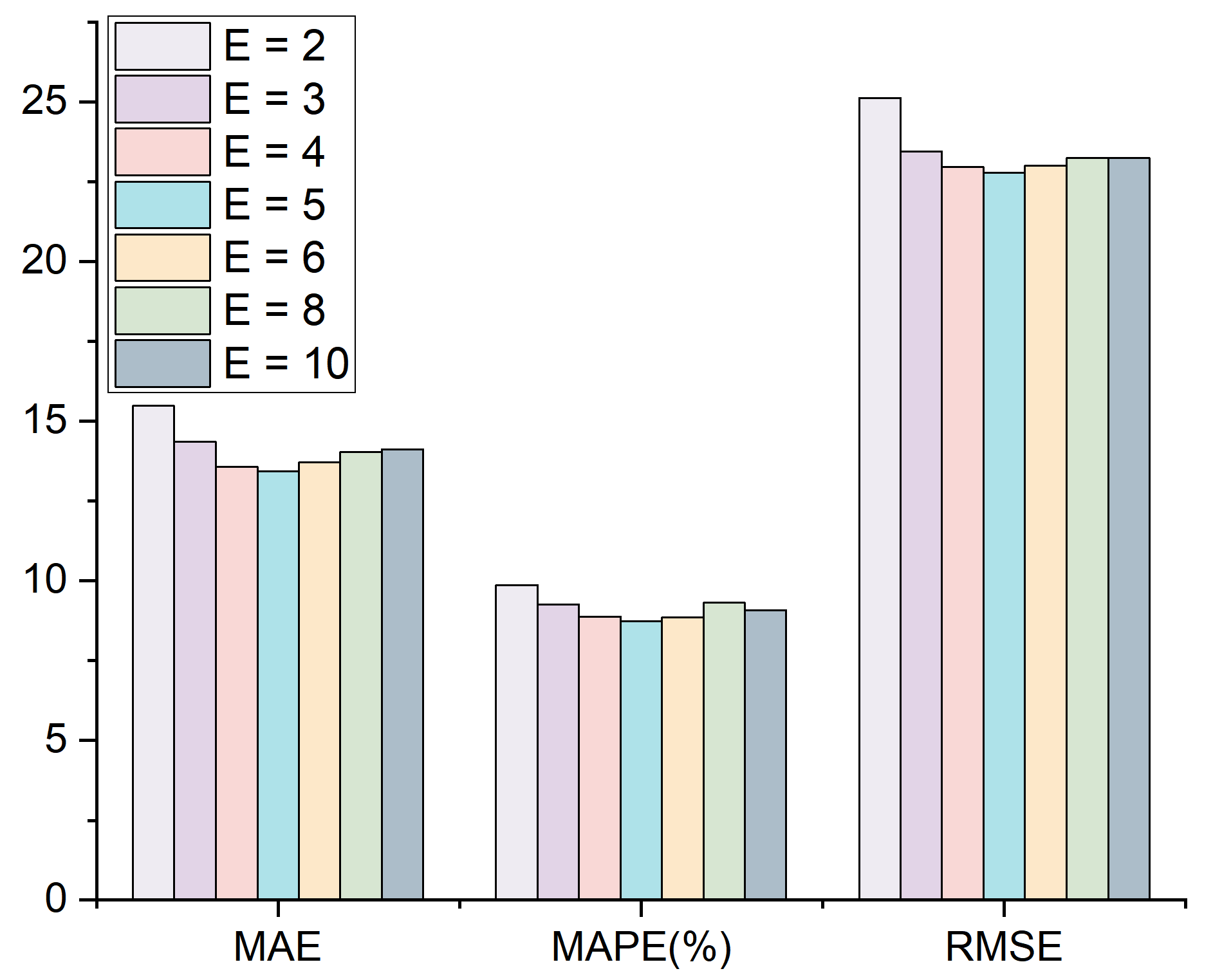}
         \caption{Sensitivity to E on PEMS08}
         \label{fig:5d}
     \end{subfigure}
     \hfill
     \caption{Impact of hyperparameter settings on  PEMS08}
        \label{fig:three graphs}
\end{figure*}
\subsection{Hyperparameters analysis of HTVGNN}
We analyzed the time-varying graph embedding dimension represented by E, the model embedding dimension represented by D, the number of enhanced temporal perception multi-head self-attention heads represented by Heads, and the number of layers of the model represented by Layers. These four hyperparameters have significant effects on PEMS08 data. The influence of the set of experimental results, \ ie, the value of D on the data set PEMS08, is 2 to 10.

\subsubsection{Sensitivity to Layers}

The impact of the model's layer count is illustrated in Figure \ref{fig:5a}. Our findings indicate that increasing the number of layers does not yield a significant improvement in performance. This observation suggests that the HTVGNN model may have reached its predictive error limit. 

\subsubsection{Sensitivity to Heads}
The impact of the number of heads on enhanced temporal perception multi-head self-attention is illustrated in Figure \ref{fig:5b}. It can be observed that the attention score calculated by the two heads can obtain a better prediction result. 
As the increase of the number of heads, spatio-temporal features will be mapped to a higher-dimensional space, resulting in more reasonable the attention score. However, an excessively large number of heads may introduce numerous negative interference and hinder accurate calculation of the attention score.

\subsubsection{Sensitivity to Dimensions}
The impact of the model embedding dimension $D$ on the model HTVGNN is illustrated in Figure \ref{fig:5c}. It can be seen that a smaller $D$ fails to effectively encode spatio-temporal features, while an excessively large $D$ may lead to overfitting.

\subsubsection{Sensitivity to E}

The impact of the size E in the embedding space for time-varying graphs is illustrated in Figure \ref{fig:5d}. From the figure, it can be observed that a smaller dimension $E$ fails to capture spatial correlations effectively, while an excessively large dimension E leads to overfitting and suboptimal learning of time-varying graphs.

\subsection{Visualiation}
\begin{figure*}
     \centering
     \begin{subfigure}[b]{0.99\textwidth}
         \centering
         \includegraphics[width=\textwidth]{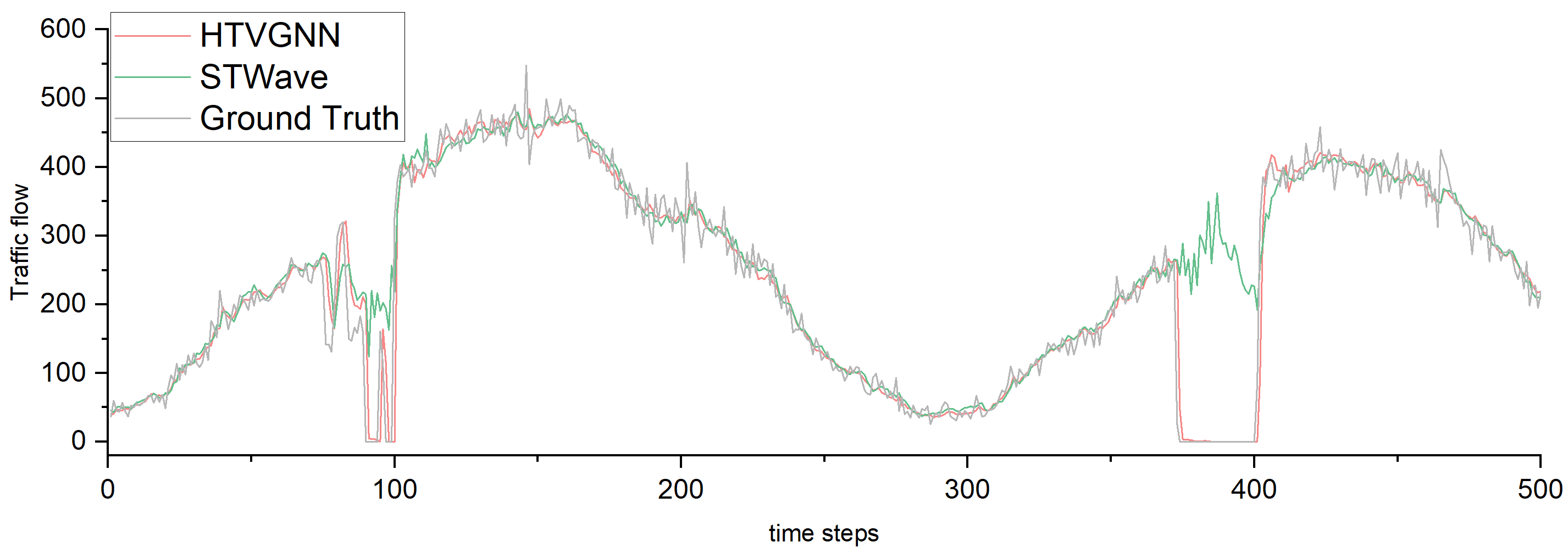}
         \caption{Node \#100 in PEMD04}
         \label{fig:y equals x}
     \end{subfigure}     \begin{subfigure}[b]{0.99\textwidth}
         \centering
         \includegraphics[width=\textwidth]{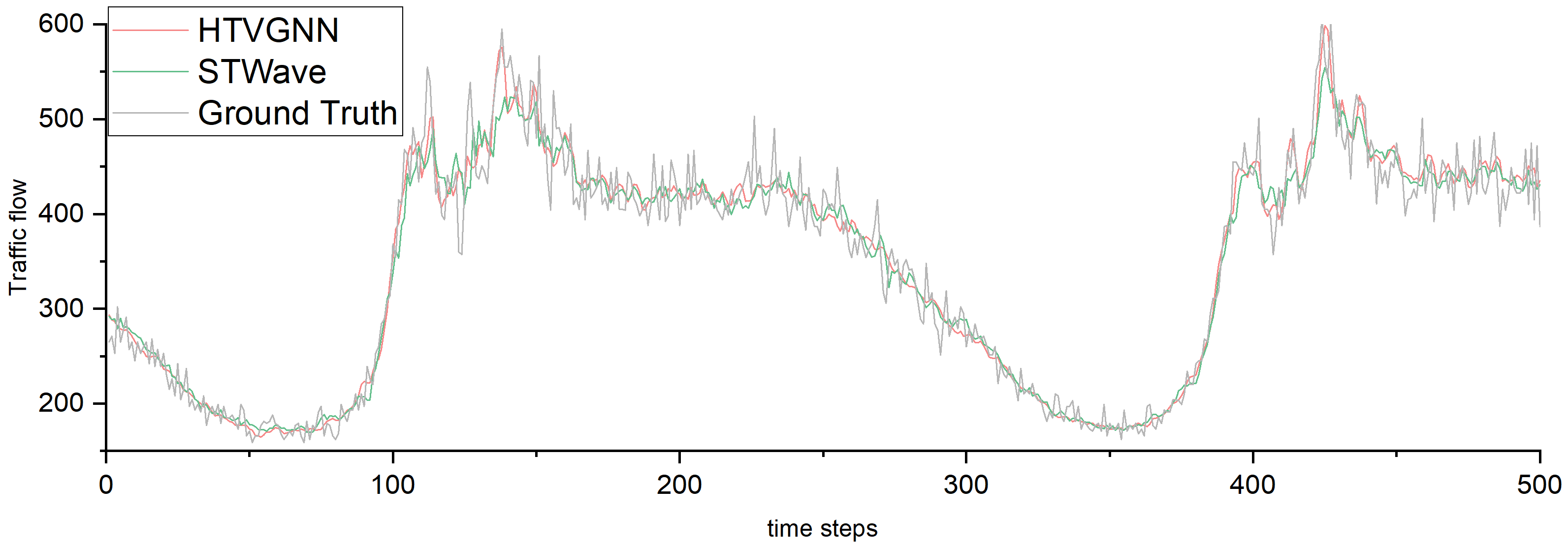}
         \caption{Node \#71 in PEMS08}
         \label{fig:y equals x}
     \end{subfigure}       \caption{Visualization results of different models on PEMS04 and PEMS08}
        \label{fg:007}
\end{figure*}

In order to visually demonstrate the superiority of our proposed model, we present the traffic flow of various benchmark models over $500$ time steps on PEMS04 and PEMS08 datasets. The comparison between traffic flow prediction results and actual values is depicted in Figure \ref{fg:007}. It can be observed that the HTVGNN model exhibits enhanced accuracy and responsiveness towards abnormal traffic flow, while also demonstrating reduced error in handling complex and dynamic traffic changes.

%
%

%
%

%
%

\subsection{Efficiency Study}
Additionally, we conducted a comparison of the training and inference efficiency of HTVGNN, Bi-STAT, and STWave on the PEMS08 dataset. To ensure fairness in our evaluation, all methods were executed on a server equipped with an Intel Core i7 13700KF processor and a single NVIDIA RTX3090 graphics card. The results of this comparison are presented in Table \ref{tb:9}.
%
%
\begin{table}[htbp]
\centering
\renewcommand\arraystretch{1.2}
\tabcolsep=0.14cm
\caption{Combination Time on the PEMS08}
\begin{tabular}{ccccccccc}     \hline
 \multicolumn{3}{c}{Model} & \multicolumn{3}{c}{Traning(sec/epochs)} & \multicolumn{3}{c}{Inference(sec/samples)}  \\ \cline{1-9}
                                             \multicolumn{3}{c}{Bi-STAT}       & \multicolumn{3}{c}{61}      & \multicolumn{3}{c}{0.0024}           \\
                     \multicolumn{3}{c}{STWave}       & \multicolumn{3}{c}{56}      & \multicolumn{3}{c}{0.0025}           \\
                          \multicolumn{3}{c}{HTVGNN}       & \multicolumn{3}{c}{50}      & \multicolumn{3}{c}{0.0061}           \\  \hline

\end{tabular}\label{tb:9}
\end{table}
%
%
As shown in Table \ref{tb:9}, despite the time-consuming nature of RNNs, the HTVGNN model utilizes a single layer. In contrast, STWave and Bi-STAT solely rely on transformer-based architectures that inherently support parallel processing. However, to capture deep spatio-temporal correlations, transformer-based models often require multiple layers. For instance, the layers of both Bi-STAT and STWave are two. Nevertheless, due to the facts that Bi-STAT includes a recollection process and STWave includes a dual-channel spatio-temporal network to model trends and events, the complexity of these two models is high.
Consequently, these transformer-only models do not offer any advantage in terms of time complexity. In comparison, although the computationally intensive RNNs was employed into our proposed one layer HTVGNN model,  HTVGNN achieves a superior prediction performance on PEMS08 dataset.
Moreover, HTVGNN significantly outperforms STWave and Bi-STAT in prediction accuracy while maintaining similar computational costs.
%
%
\section{Conclusion}
In this paper, we propose a novel hybrid time-varying graph neural network prediction method, HTVGNN. To accurately capture the most relevant temporal features in temporal correlation modeling, we introduce an enhanced temporal perception multi-head self-attention mechanism based on time-varying mask enhancement. We allocate the optimal attention calculation scheme for different temporal features. In spatial correlation modeling, we employ two strategies. Firstly, in static spatial correlation modeling, we design a coupled graph learning mechanism to integrate the graph learned at each time step and effectively enhance the model's long-term prediction capability. Secondly, in dynamic spatial correlation modeling within the road network, we define a mask based on topological matrix and semantic matrix to optimize the dynamic time-varying graph and improve the model's ability to capture dynamic spatial correlations. Simulation experiments conducted on four real datasets validate our proposed method HTVGNN by demonstrating its superior prediction accuracy compared with state-of-the-art spatio-temporal graph neural network models. Furthermore, our proposed coupling graph learning method significantly improves long-term prediction performance of the model as observed in ablation experiments. Future work will focus on exploring more effective masking techniques through model pre-training and providing more efficient yet powerful masking mechanisms for Transformer models.

%
%

\appendices
%
%
%


%
\bibliographystyle{IEEEtran}  
\bibliography{refbib}     
\begin{IEEEbiography}[{\includegraphics[width=1in,height=1.25in,clip,keepaspectratio]{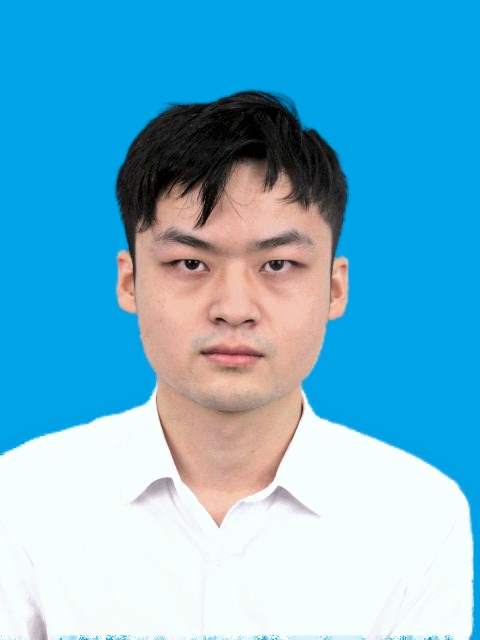}}]{Ben-Ao Dai} received the B.S. degree in Communication Engineering from the Zhejiang Wanli University, Ningbo, Zhejiang, China, in 2020. He is currently pursuing the M.S. degree in Electronic Information at Zhejiang Sci-tech University, Xiasha Campus, Hangzhou, Zhejiang, China. He conducts a two-year research project with his tutor in the School of Information Science and Engineering, Jiaxing University, Jiaxing 314001, China, during his study for a 
master's degree. His research interests include spatial-temporal data predictions, artificial intelligence.
\end{IEEEbiography}
\begin{IEEEbiography}[{\includegraphics[width=1in,height=1.25in,clip,keepaspectratio]{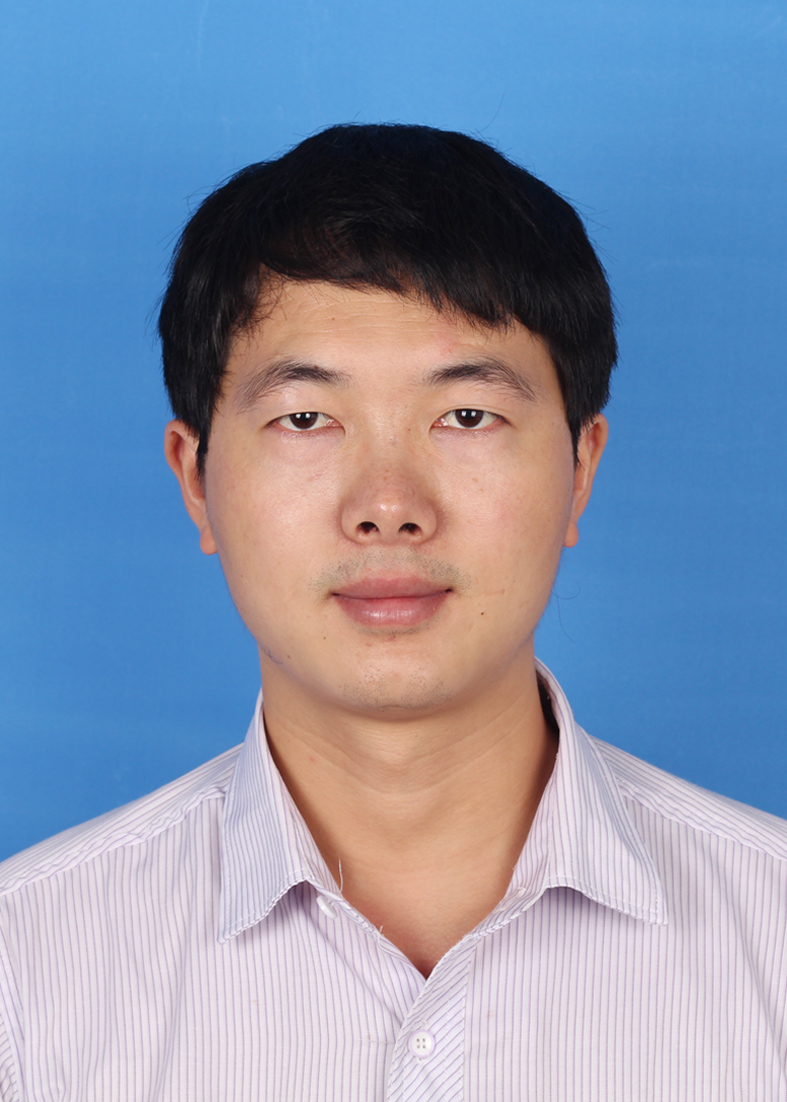}}]{Bao-Lin Ye} is currently an associate professor in the Department of Electronic and Information Engineering at School of Information Science and Engineering, Jiaxing University, Jiaxing, Zhejiang, China. He is also a tutor of Professional Master's Degree in Electronic Information at Zhejiang Sci-tech University, Hangzhou, Zhejiang, China. He received the Ph.D. degree from Zhejiang University, Hangzhou, China, in June 2015. He was a Visiting Scholar at Indiana University-Purdue University Indianapolis, Indianapolis, IN, USA, from 2018 to 2019. His research interests include deep learning, reinforcement learning, intelligent control theory, and autonomous vehicle control.
\end{IEEEbiography}
\begin{IEEEbiography}[{\includegraphics[width=1in,height=1.25in,clip,keepaspectratio]{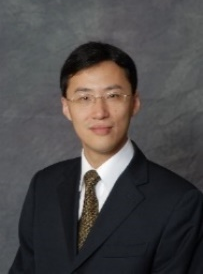}}]{Lingxi Li}
(Senior Member, IEEE) 
is currently a professor in the Elmore Family School of Electrical and Computer Engineering, Purdue University, Indianapolis, IN 46202, USA. He received his PhD degree in electrical and computer engineering from the University of Illinois at Urbana-Champaign in 2008. He has authored/co-authored one book and over 150 research articles in refereed journals and conferences. He is currently serving as an associate editor for five international journals and has served as general chair, program chair, program co-chair, publication chair, etc., for more than 30 international conferences. His current research focuses on the modeling, analysis, control, and optimization of complex systems, connected and automated vehicles, intelligent transportation systems, discrete event dynamic systems, human-machine interaction, and parallel intelligence.
\end{IEEEbiography}

\end{document}